\documentclass[12pt, fullpage]{article}

\usepackage{amssymb}
\usepackage{amsmath}
\usepackage{appendix}
\usepackage{verbatim}
\usepackage{graphicx}
\usepackage{multirow}
\usepackage{caption}
\usepackage{subfigure}
\usepackage{multicol}
\usepackage{algorithmic}
\usepackage{algorithm}
\usepackage{url}

\def\be{\begin{eqnarray}}
\def\bse{\begin{eqnarray*}}
\def\ee{\end{eqnarray}}
\def\ese{\end{eqnarray*}}
\def\bee{\begin{enumerate}}
\def\eee{\end{enumerate}}
\def\bed{\begin{description}}
\def\eed{\end{description}}
\def\bei{\begin{itemize}}
\def\eei{\end{itemize}}

\def\wt{\widetilde}
\def\wh{\widehat}

\def\bfa{{\bf a}}  \def\bfA{{\bf A}}
\def\bfb{{\bf b}}  \def\bfB{{\bf B}}
  
  \def\bfD{{\bf D}}

  \def\bfI{{\bf I}}

  \def\bfM{{\bf M}}

\def\bfp{{\bf p}}  \def\bfP{{\bf P}}
\def\bfq{{\bf q}}  \def\bfQ{{\bf Q}}
  \def\bfR{{\bf R}}

\def\bfx{{\bf x}}  
\def\bfy{{\bf y}}  
  
\def\boldsigma{{\mbox{\boldmath $\sigma$}}}

\def\T{{\mbox{\rm\tiny T}}}

\usepackage{url} 
\usepackage{float}
\usepackage{graphicx}
\usepackage{amsfonts, amssymb, amscd}
\usepackage{amsmath, delarray, theorem}
\usepackage[numbers, sort]{natbib}
\renewcommand{\cite}{\citep}

\newcommand{\bitem}{\begin{itemize}}
\newcommand{\eitem}{\end{itemize}}
\newcommand{\benum}{\begin{enumerate}}
\newcommand{\eenum}{\end{enumerate}}
\newcommand{\beqnn}{\begin{eqnarray*}}
\newcommand{\eeqnn}{\end{eqnarray*}}
\newcommand{\beqn}{\begin{eqnarray}}
\newcommand{\eeqn}{\end{eqnarray}}
\newcommand{\normal}[2]{\mbox{N}(#1,#2)}

\usepackage{mywidefile}

\begin{document}

\title{Content-boosted Matrix Factorization Techniques for Recommender Systems}
\author{Jennifer Nguyen \\
Computer Science Department \\
University College London \\
London, England WC1E 6BT \\
\\
Mu Zhu \\
Department of Statistics and Actuarial Science \\
University of Waterloo\\ 
Waterloo, Ontario, Canada N2L 3G1}
\date{\today}
\maketitle

\begin{abstract} Many businesses are using recommender systems for 
marketing outreach. Recommendation algorithms can be either based on 
content or driven by collaborative filtering. We study different ways to 
incorporate content information directly into the matrix factorization 
approach of collaborative filtering. These content-boosted matrix 
factorization algorithms not only improve recommendation accuracy, but 
also provide useful insights about the contents, as well as make 
recommendations more easily interpretable. \end{abstract}

\vspace{3mm}

\section{Introduction} 
\label{sec:intro}

Many businesses today are using the Internet to promote and to sell 
their products and services. Through the Internet, businesses can easily 
market many items to a large number of consumers. With a vast number of 
items, however, consumers may be overwhelmed by their choices. That is 
why, in an effort to maintain customer satisfaction and loyalty, many 
businesses have also integrated the use of recommender systems in their 
marketing strategies. For example, the online store \url{www.amazon.com} 
will suggest, based on a user's past purchases, products that he or she 
may be interested in.

Recommender systems today typically use one of two approaches: the 
\textit{content-based} approach, or the \textit{collaborative filtering} 
(CF) approach. In the content-based approach (e.g., Pandora, 
\url{www.pandora.com}), a profile is created for each user and for each 
item. The user profile describes the contents that he or she likes, and 
the item profile describes the contents that it contains. To a given user, 
the system recommends items that match his or her profile. In the CF 
approach (e.g., Netflix, \url{www.netflix.com}), users who have rated the 
same items closely are considered to have similar preferences overall. To 
a given user, the system recommends items that similar users have rated 
favorably before.

For an extensive review and discussion of different CF algorithms as well 
as an up-to-date and comprehensive bibliography, we refer the readers 
to a recent article by \citet{netflix-review-statsci}. While 
various algorithms have been adapted for the recommendation problem 
including restricted Boltzmann machines \citep{cf-rbm}, most CF algorithms 
can be classified into two broad categories \cite{netflix-nbr}: those 
based on \textit{nearest neighbors} and those based on \textit{matrix 
factorization}. While the nearest-neighbor approach is more intuitive, the 
matrix-factorization approach has gained popularity as a result of the 
Netflix contest \cite{Koren09}. 

\subsection{Focus of paper}
\label{sec:limitedcomparison}

Perhaps the most important lessons from the Netflix contest are that, in 
terms of prediction accuracy, it is often difficult for any single 
algorithm to outperform an ensemble of many different algorithms 
\citep{Koren09, netflix-review-statsci}, and that algorithms of different 
flavors, when taken alone, often have similar predictive power for a 
given problem, even though they may each capture different effects. 

For example, as shown by \citet{netflix-review-statsci} in their Table~1, 
on the Netflix data, a neighborhood-based method alone (labelled ``kNN'' 
in their table) had a root mean squared error (RMSE) of $0.9174$ whereas a 
method based on matrix factorization alone (labelled ``SVD'' in their 
table) had an RMSE of $0.9167$ --- very close indeed. A significant drop 
in the RMSE (to $0.8982$) was achievable only when the two classes of 
methods were combined together; see also \citet{netflix-nbr}. And it is 
widely known that the ultimate winner in the Netflix contest (with an RMSE 
of $0.8572$) was an ensemble of no fewer than 800 different algorithms.

Therefore, a research project on CF can either focus on new classes of CF 
algorithms that are fundamentally different from existing ones, or focus 
on improvements or extensions {\em within} a certain class. For projects 
of the first type, the key question is whether the new class of algorithms 
is better than other classes. For those of the second type, the key 
question is whether the proposed extension adds any value when compared 
with baseline algorithms in the {\em same} class. The research we will 
report in this paper is strictly of the second type. In particular, we 
focus on the matrix factorization approach only.

\subsection{The ``cold start'' problem}
\label{sec:coldstart}

One advantage of the CF approach is that it does not require extra
information on the users or the items; thus, it is capable of recommending 
an item without understanding the item itself \cite{survey-CF}. However, 
this very advantage is also the root cause of the so-called ``cold start" 
problem, which refers to the general difficulty in performing CF for users 
and items that are relatively new. By definition, newer users are those 
who have not rated many items, so it is difficult to find other users with 
similar preferences. Likewise, newer items are those which have not been 
rated by many users, so it is difficult to recommend them to anyone.

Various ideas have been proposed to deal with the ``cold start'' problem. 
\citet{Park06} suggested using so-called ``filterbots'' --- artificial 
items or users inserted into the system with pre-defined characteristics. 
For instance, an action-movie filterbot can make recommendations to new 
users who have only liked one or two action movies. More recently, 
\citet{Zhao11} suggested \textit{shared} CF, an ensemble technique that 
aggregates predictions from several different systems. Since one 
recommender system may have data on user-item pairs that another one does 
not, it is possible to improve recommendations by sharing information 
across different systems.

Another common approach for dealing with the ``cold start'' problem is to 
fill in the missing ratings with ``pseudo'' ratings before applying CF. 
For example, \citet{Goldberg01} did this with principal component 
analysis; \citet{Nguyen07} did this with rule-based induction; while 
\citet{Melville02} did this with a hybrid, two-step approach, creating 
``pseudo'' ratings with a content-based classifier.

\subsection{Objectives and contributions}
\label{sec:goal}

The key idea behind the hybrid approach is to leverage supplemental 
information \cite{Semeraro09}. Many recent works have taken this basic 
idea to new heights, successfully exploiting supplemental information from 
different sources and in various forms, for example, tagging history 
\cite{cf-YZ}, personality traits \cite{Nunes09, Hu11}, social networks 
\cite{Jamali10, Yu11}, and Wikipedia articles \cite{Katz11}.

In this paper, we focus on a particular type of supplemental information 
--- content information about the individual items. For example, for 
recipes \cite{Forbes11} we may know their ingredient lists; for movies 
\cite{Koren09} we may know their genres. Moreover, we focus on ways to 
take advantage of such content information {\em directly} in the matrix 
factorization approach, not by using a hybrid or two-step algorithm. We 
refer to our suite of algorithms as ``content-boosted matrix factorization 
algorithms''.

Not only can these content-boosted algorithms achieve improved 
recommendation accuracy (Section~\ref{sec:rslt}), they can also produce 
more interpretable recommendations (Section~\ref{sec:CB2-rslt}), as well 
as furnish useful insights about the contents themselves that are 
otherwise unavailable (Section~\ref{sec:CB1-rslt}). More interpretable 
recommendations are becoming ever more desirable commercially, because 
users are more likely to act on a recommendation if they understand why it 
is being made to them \cite{recsys11-ebay, rec-transparency}, while better 
understandings of contents can facilitate the creation of new products, 
such as recipes with substitute ingredients.

One of the big lessons from the Netflix contest was the need to develop 
``diverse models'' that capture ``distinct effects, even if they're very 
small effects'' \cite{netflix-wired-news}. Since most methods do not take 
content information into account, the value of our content-boosted 
algorithms is clear, and we fully expect that our algorithms will further 
enhance any existing ensembles.

\subsection{Outline}

We proceed as follow. In Section~\ref{sec:BL}, we give a brief review of 
the matrix factorization (MF) approach for collaborative filtering. In 
Section~\ref{sec:main}, we present a number of different content-boosted 
MF algorithms. In Section~\ref{sec:expr}, we describe the data sets we 
used and the experiments we performed to study and evaluate various 
algorithms. In Section~\ref{sec:disc}, we describe useful by-products from 
these content-boosted MF techniques. In Section~\ref{sec:future}, we 
briefly mention some open problems for future research. We end in 
Section~\ref{sec:summary} with a brief summary.

\section{Matrix factorization: A brief review}
\label{sec:BL}
\def\grad{\nabla}

Before we start, it is necessary to review the basic matrix factorization 
method briefly. Our review follows the work of \citet{Koren09}.

\subsection{Notation}

Given a set of users 
$U=\{u_1,\ldots,u_N\}$, and a set of items $I=\{i_1,\ldots,i_M\}$, let 
$r_{ui}$ denote the rating given by user $u$ to item $i$. These ratings 
form a user-item rating matrix, $\bfR=[r_{ui}]_{N\times M}$. In principle, 
$r_{ui}$ can take on any real value but, in practice, $r_{ui}$ is 
typically binary, indicating ``like'' and ``dislike'', or integer-valued 
in a certain range, indicating different levels of preferences, e.g., 
$r_{ui} \in \{1,\ldots,5\}$.

Often, the rating matrix $\bfR$ is highly sparse with many unknown 
entries, as users typically are only able to rate a small fraction of the 
items --- recall the ``cold start'' problem discussed briefly in 
Section~\ref{sec:coldstart}. We denote $$T=\{(u,i): r_{ui} \text{ is 
known}\}$$ as the set of indices for known ratings. Given an unknown 
(user, item)-pair, $(u,i)\not\in T$, the goal of the recommender system is 
to predict the rating that user $u$ would give to item $i$, which we 
denote by $\wh r_{ui}$. Furthermore, we define $$T_{u\cdot} \equiv \{i: 
(u,i) \in T\}$$ to be the set of items that have been rated by user $u$, 
and $$T_{\cdot i} \equiv \{u: (u,i) \in T\}$$ to be the set of users who 
have rated item $i$.

\subsection{Normalization by ANOVA}
\label{sec:anova}

Despite its overwhelming simplicity, an ANOVA-type of model often captures 
a fair amount of information in the rating data \citep{Koren09, 
netflix-review-statsci}. The simplest ANOVA-type model used in the 
literature consists of just main effects, i.e.,
\beqn
\label{eq:anova}
r_{ui} = \mu + \alpha_u + \beta_i + \epsilon_{ui},
\eeqn
where $\epsilon_{ui}$ is white noise, $\mu$ is the overall mean, 
$\alpha_u$ represents a user-effect, and 
$\beta_i$ represents an item-effect. These two main effects capture the 
obvious fact that some items are simply better liked than others, while 
some users are simply more difficult to please.

It is common in the literature to normalize the rating matrix 
$\bfR$ by removing such an ANOVA-type model 
before applying any matrix-factorization (or nearest-neighbor) methods 
\citep[e.g.,][]{Koren09}. In all of our experiments reported below, 
we followed this common practice, that is, all matrix-factorization 
algorithms were applied to $r_{ui} - \wh\mu - \wh\alpha_u - \wh\beta_i$, 
and the predicted rating was actually $\wh r_{ui} + \wh\mu + \wh\alpha_u 
+ \wh\beta_i$, where $\wh r_{ui}$ was the prediction from the 
matrix-factorization algorithm, and $\wh \mu$, $\wh \alpha_u$, $\wh 
\beta_i$ were the MLEs of $\mu$, $\alpha_u$, $\beta_i$. In order not to 
further complicate our notation, however, this detail will 
be suppressed in our presentation, and we still 
use the notations, $r_{ui}$ and $\bfR$, despite the normalization step.

\subsection{Matrix factorization}
\label{sec:BLdetails}
\def\BL{{\mbox{\rm\tiny BL}}}

To predict unknown ratings in $\bfR$, the matrix factorization approach
uses all the known ratings to decompose the matrix $\bfR$ into the 
product of two low-rank, latent feature matrices, one for the users, 
$\bfP_{N \times K}$, and another for the items, $\bfQ_{M \times K}$, 
so that 
\begin{eqnarray}
\bfR \approx \wh\bfR = \bfP\bfQ^{\T} = 
\underbrace{\left[\begin{array}{c} 
\bfp_1^{\T} \\ 
\bfp_2^{\T} \\ 
\vdots \\
\bfp_N^{\T} \\ 
\end{array}\right]}_{N \times K}
\underbrace{\left[\begin{array}{cccc}
\bfq_1 & \bfq_2 & \cdots & \bfq_M
\end{array}\right]}_{K \times M}. \label{product}
\end{eqnarray}
The 
latent feature vectors --- $\bfp_{u}$ for user $u$ ($u=1,2,...,N$) and 
$\bfq_{i}$ for item $i$ ($i=1,2,...,M$) --- are $K$-dimensional, 
where $K \ll \min\{M,N\}$ is pre-specified. The predicted rating for the 
user-item pair 
$(u,i)$ is simply $$\wh r_{ui} = \bfp_{u}^{\T}\bfq_{i}.$$
Intuitively, one can imagine a $K$-dimensional map, in which
$\bfp_u$ and $\bfq_i$ are the (latent) coordinates for 
user $u$ and item $i$, respectively, and
all the information that we need in order to make recommendations is 
contained in such a map --- users will generally like items that 
are nearby.
Latent-coordinate models have a long history, e.g., principal component 
analysis, factor analysis, multidimensional scaling, and so on 
\citep[see, e.g.,][]{mardia}.

Mathematically, the factorization (\ref{product}) can be achieved by 
solving the 
optimization problem,
\begin{eqnarray}
\label{basicMF}
\underset{\bfP,\bfQ}{\min} \quad 
\| \bfR - \bfP\bfQ^{\T} \|^2,
\end{eqnarray}
where $\| \cdot \|$ is the Frobenius norm.
To prevent over-fitting, it is common to include a regularization penalty 
on the sizes of $\bfP$ and $\bfQ$, turning the optimization problem 
above into
\begin{eqnarray}
\underset{\bfP,\bfQ}{\min} \quad
\| \bfR - \bfP\bfQ^{\T} \|^2 + 
\lambda\left( \| \bfP \|^2 + \| \bfQ \|^2 \right).
\label{obj-raw}
\end{eqnarray}
From a Bayesian point of view, the first part of the objective 
function (\ref{obj-raw}) can be viewed as coming from a Gaussian 
likelihood function; 
the regularization penalties can be viewed as coming from spherical 
Gaussian priors on the user and item feature vectors; and the solution to 
the optimization problem itself is then the so-called maximum a posteriori 
(MAP) estimate \citep{MFprob}. 

\subsection{Relative scaling of penalty terms}
\label{sec:gamma}

\citet{netflix-review-statsci} used empirical Bayes analysis to argue that 
one should, in principle, always penalize $\|\bfp_u\|^2$ and 
$\|\bfq_i\|^2$ by different amounts. In practice, their advice is not 
always followed because the extra computational burden to select two 
tuning parameters rather than one is substantial, and the resulting 
payoff in terms of performance improvement may not be significant.

In our work, we found it convenient to scale the second penalty term --- 
the one on $\|\bfQ\|^2$ --- by a factor $\gamma>0$ such  
that, regardless of how many users ($N$) and how many items ($M$) there 
are, the penalty on $\|\bfQ\|^2$ is always on the same order of 
magnitude as the penalty on $\|\bfP\|^2$. We will come back to 
this point later (Section~\ref{sec:gamma-more}).

Furthermore, since most entries in $\bfR$ are unknown, we can only 
evaluate the first term in (\ref{obj-raw}) over known entries $(u,i) \in 
T$. This means the optimization problem actually solved in practice is:
\begin{eqnarray}
\underset{\bfP,\bfQ}{\min} \quad
L_{\BL}(\bfP,\bfQ) = 
\sum_{(u,i)\in T} (r_{ui} - \bfp_{u}^{\T}\bfq_{i})^2 
+ \lambda\left( \sum_{u} \|\bfp_u \|^2 + 
                \gamma \sum_{i} \|\bfq_i \|^2
         \right).
\label{obj-BL}
\end{eqnarray}
The subscript ``BL'' stands for ``baseline''.
For the purpose of comparison, we will refer to this method below as the 
baseline matrix factorization method, or simply the baseline (BL)
algorithm.

\subsection{Alternating gradient descent}

With both $\bfP$ and $\bfQ$ being unknown, the optimization problem 
(\ref{obj-BL}) is not convex. It can be solved using an alternating 
gradient descent algorithm \cite{Koren09}, moving along the gradient with 
respect to $\bfp_u$ while keeping $\bfq_i$ fixed, and vice versa.

Let $\grad^{\BL}_u$ denote the derivative of $L_{\BL}$ with respect to 
$\bfp_u$ 
and $\grad^{\BL}_i$, its derivative with respect to $\bfq_i$.
Then, 
\begin{eqnarray}
\grad^{\BL}_u &\propto& 
\sum_{i \in T_{u\cdot}} -(r_{ui} - \bfp_u^{\T}\bfq_i)\bfq_i + \lambda\bfp_u,
\label{dQu} \\
\grad^{\BL}_i &\propto&
\sum_{u \in T_{\cdot i}} -(r_{ui} - \bfp_u^{\T}\bfq_i)\bfp_u + \lambda\gamma\bfq_i,
\label{dQv}
\end{eqnarray}
for every $u=1,2,...,N$ and $i=1,2,...,M$.
At iteration $(j+1)$, the updating equations for $\bfp_u$ and $\bfq_i$ 
are:
\begin{eqnarray}
\bfp_u^{(j+1)} &=& \bfp_u^{(j)} -
 \eta
 \grad^{\BL}_u\left(\bfp_u^{(j)},\bfq_i^{(j)}\right), \label{newU}\\
\bfq_i^{(j+1)} &=& \bfq_i^{(j)} - 
 \eta
 \grad^{\BL}_i\left(\bfp_u^{(j)},\bfq_i^{(j)}\right), \label{newV}
\end{eqnarray}
where $\eta$ is the step size or learning rate. The algorithm is typically 
initialized with small random entries for $\bfp_u$ and $\bfq_i$, and 
iteratively updated over all $u=1,\ldots,N$ and $i=1,\ldots,M$ until 
convergence (see Algorithm~\ref{alg1}). We will say more about 
initialization later (Section~\ref{sec:init}).

\renewcommand{\algorithmicrequire}{\textbf{Input:}}
\renewcommand{\algorithmicensure}{\textbf{Output:}}
\begin{algorithm}
\caption{Alternating Gradient Descent Algorithm for Optimizing $L_{\BL}$ 
--- Eq.~(\ref{obj-BL})}
\label{alg1}
\begin{algorithmic}[1]
\REQUIRE $\bfR=[r_{ui}]_{N \times M}$, 
$K$
\ENSURE $\bfP$, $\bfQ$
\STATE \textbf{initialize} 
$j\gets 0$ and
choose $\bfP^{(0)},\bfQ^{(0)}$ (see Section~\ref{sec:init})
\REPEAT
	\FORALL{$u=1,\ldots,N$ and $i=1,\ldots,M$}
		\STATE compute 
                        $\grad^{\BL}_u$ and 
                        $\grad^{\BL}_i$ using (\ref{dQu})-(\ref{dQv})
		\STATE update $\bfp_u^{(j+1)}$ and
			$\bfq_i^{(j+1)}$ with (\ref{newU})-(\ref{newV})
	\ENDFOR
\UNTIL{$[L_{\BL}(\bfP^{(j)},\bfQ^{(j)})-
         L_{\BL}(\bfP^{(j+1)},\bfQ^{(j+1)})]/
         L_{\BL}(\bfP^{(j)},\bfQ^{(j)})<\varepsilon$}
\RETURN $\bfP,\bfQ$
\end{algorithmic}
\end{algorithm}

\subsection{SVD and other matrix factorization techniques}
\label{sec:svd}

In the CF literature, the matrix factorization approach outlined above is 
often dubbed the ``singular value decomposition (SVD) approach'' 
\citep[see, e.g.,][]{funk, Koren09, netflix-review-statsci}. 
Strictly speaking, this is a bit misleading. The SVD is perhaps the single 
most widely used matrix factorization technique in all of applied mathematics; 
it solves the following problem:
\beqn
\min &\quad& \| \bfR - \bfP_* \bfD_* \bfQ_*^{\T} \|^2 \label{eq:SVD} \\
\mbox{s.t.} 
&\quad& \bfD_* \mbox{ is diagonal with rank $K$}, \notag \\
&\quad& \bfP_*^{\T}\bfP_*=\bfI \quad\mbox{and}\quad 
        \bfQ_*^{\T}\bfQ_*=\bfI. \notag
\eeqn
By letting
$\bfP=\bfP_*\bfD^{1/2}$ and
$\bfQ=\bfQ_*\bfD^{1/2}$, SVD would give us
\beqn
\label{eq:SVD-rslt}
 \bfR \approx \bfP \bfQ^{\T}
\eeqn
such that $\bfP^{\T}\bfP=\bfD_*^{1/2}\bfP_*^{\T}\bfP_*\bfD_*^{1/2}=\bfD_*$ 
is diagonal, meaning that $\bfP$ is an orthogonal matrix, and likewise for 
$\bfQ$. However, the matrix factorization approach outlined above does 
{\em not} require either $\bfP$ or $\bfQ$ to be orthogonal. To be sure, we 
confirmed this directly with the winners of the Netflix contest 
\citep{svd-volinsky}, who used this technique pervasively in their work. 
Without the orthogonality constraints, this would certainly raise 
identifiability and degeneracy questions for the optimization problem 
(\ref{obj-BL}), but these problems can be avoided {\em in practice} by 
carefully initializing the alternating gradient descent algorithm --- we 
elaborate on this detail in Section~\ref{sec:init} below.

\citet{nmf-leeseung} popularized 
another matrix factorization technique 
called the non-negative matrix factorization (NMF), which is 
(\ref{basicMF}) 
with the additional non-negativity constraints that
\[
P_{uk} \geq 0 \quad\mbox{and}\quad 
Q_{ik} \geq 0 \quad\mbox{for all}\quad u,i,k.
\]
The NMF has been used to analyze a wide variety of data such as images 
\citep{nmf-leeseung} and gene expressions \citep{nmf-metagene} to reveal 
interesting underlying structure. In recent years, it has also been used 
to perform CF \citep[e.g.,][]{cf-YZ-NMF, ensembleMF} even though finding 
underlying structures in the data is often not the primary goal for CF. 
Matrix factorization with either orthogonality constraints (e.g., SVD) or 
nonnegativity constraints (e.g., NMF) is more sound mathematically, since 
the problem is somewhat ill-defined without any constraints. However, we 
will still focus only on the unconstrained version outlined above 
(Section~\ref{sec:BLdetails}) since it remains the most dominant in the CF 
community, owing partly to its wide use in the three-year-long Netflix 
contest.

\section{Content-boosted matrix factorization}
\label{sec:main}
\def\wt{\widetilde}

Now, suppose that, for each item $i$, there is a content vector $\bfa_i = 
[a_{i1},\ldots, a_{iD}]$ of $D$ attributes. Stacking these vectors 
together gives an attribute matrix, $\bfA=[a_{id}]_{M \times D}$. For 
simplicity, we assume that all entries in $\bfA$ are binary, i.e., $a_{id} 
\in \{0,1\}$, each indicating whether item $i$ possesses attribute $d$. In 
what follows, we study and compare different ways of incorporating this 
type of content information {\em directly} into the matrix factorization 
approach. We present two classes of methods with slightly different 
flavors. One class uses extra penalties with selective shrinkage effects 
(Section~\ref{sec:CB2}), and the other uses direct regression constraints 
(Section~\ref{sec:CB1}).

\subsection{Alignment-biased factorization}
\label{sec:CB2}
\def\AB{{\mbox{\rm\tiny AB}}}
\def\gAB{{\mbox{\rm\tiny gAB}}}

To incorporate $\bfA$ into the matrix factorization approach, one idea is 
as follows: if two items $i$ and $i'$ share at least $c$ attributes in 
common --- call this the ``common attributes'' condition, then it makes 
intuitive sense to require that their feature vectors, $\bfq_i$ and 
$\bfq_{i'}$, be ``close'' in the latent space.

\subsubsection{Details}
\label{sec:CB2-details}

For the matrix factorization approach, it is clear from (\ref{product}) 
that the notion of closeness is modeled mathematically by the inner 
product in the latent feature space. Therefore, to say that $\bfq_i$ and 
$\bfq_{i'}$ are ``close'' means that their inner product, 
$\bfq_i^{\T}\bfq_{i'}$, is large. We can incorporate this preference by 
adding another penalty, which we call the ``alignment penalty'', to 
the optimization problem (\ref{obj-BL}).

For binary $a_{id}$, the ``common attributes'' condition is easily 
expressed by $\bfa_i^{\T} \bfa_{i'} \geq c$. 
Let 
$$\mathcal{S}_c(i) \equiv \{i': \quad i' \neq i 
 \quad\mbox{and\quad} \bfa_i^{\T}\bfa_{i'} \geq c\}.$$
We solve the following 
optimization problem:
\begin{eqnarray}
\underset{\bfP,\bfQ}{\min} \quad L_{\AB}(\bfP,\bfQ) = 
 L_{\BL}(\bfP,\bfQ) - \underbrace{\lambda\gamma
 \sum_{i=1}^M
 \sum_{i' \in \mathcal{S}_c(i)}
 \frac{\bfq_i^{\T}\bfq_{i'}}{|\mathcal{S}_c(i)|}
 }_{\mbox{alignment penalty}},
\label{obj-CB2}
\end{eqnarray}
where $L_{\BL}(\bfP,\bfQ)$ is the baseline objective function given by 
(\ref{obj-BL}), and the 
notation $|\mathcal{S}|$ means the size of the set $\mathcal{S}$.
Notice that we make the alignment penalty adaptive to the size of 
$\mathcal{S}_c(i)$. The subscript ``AB'' stands for ``alignment-biased''.

It is easy to see that the basic idea of alternating gradient descent 
still applies. 
For $L_{\AB}$, the gradient with respect to 
$\bfp_u$ clearly remains the same, that is,
\[
 \grad^{\AB}_u = \grad^{\BL}_u,
\]
while the gradient with respect to $\bfq_i$ becomes
\begin{eqnarray}
\grad^{\AB}_i &\propto&
\sum_{u \in T_{\cdot i}} -(r_{ui} - \bfp_u^{\T}\bfq_i)\bfp_u + \lambda\gamma
 \left[\bfq_i - \sum_{i'\in \mathcal{S}_c(i)} 
 \frac{\bfq_{i'}}
      {|\mathcal{S}_c(i)|}
 \right].
\label{dQv2}
\end{eqnarray}
The 
updating equations are identical to (\ref{newU})-(\ref{newV}), except
that
$\grad^{\BL}_u$ and $\grad^{\BL}_i$ are replaced by
$\grad^{\AB}_u$ and $\grad^{\AB}_i$.

\subsubsection{Differential shrinkage effects}
\label{sec:diffshrink}

The effect of the alignment penalty can be seen explicitly from 
(\ref{dQv2}) as shrinking the latent vector of each item toward the 
centroid of items that share a certain number of attributes with it. This 
is the selective shrinkage effect that we alluded to earlier 
(Section~\ref{sec:main}, page~\pageref{sec:main}), and it plays a central 
role.

Next, we introduce a generalized/smoothed version of our alignment penalty 
(Section~\ref{sec:smooth}) as well as a related but slightly different 
mathematical formulation (Section~\ref{sec:tag}). We will see that the 
main difference between these methods lies in their respective shrinkage 
effects --- in each iteration, they shrink $\bfq_i$ towards slightly 
different centroids and by slightly different amounts; see the terms 
inside the square brackets in (\ref{dQv2}), (\ref{dQv2smooth}), 
(\ref{dQv3}) and (\ref{dQv3-easy}).

\subsubsection{A smooth generalization}
\label{sec:smooth}

\begin{figure}[ht]
\centering
\includegraphics[width=0.5\textwidth, angle=270]{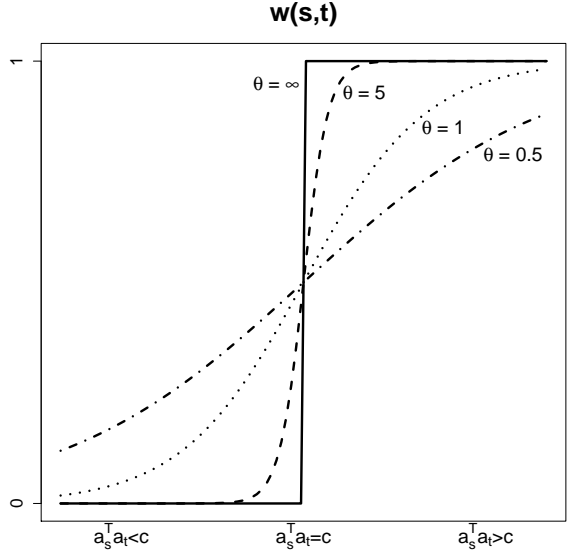}
\caption{The function $w(s,t)$ as given by (\ref{softweight}), for 
$\theta = 0.5, 1, 5, \infty$.}\label{showstep}
\end{figure}

An obvious generalization of the alignment penalty is to change 
(\ref{obj-CB2}) into
\begin{eqnarray}
\underset{\bfP,\bfQ}{\min} \quad L_{\gAB}(\bfP,\bfQ) = 
 L_{\BL}(\bfP,\bfQ) - \underbrace{\lambda\gamma
 \sum_{i=1}^M\sum_{i'=1}^M w(i,i') \bfq_i^{\T} \bfq_{i'}
 }_{\mbox{gen.~alignment penal.}},
\label{obj-gAB}
\end{eqnarray}
with
\begin{eqnarray}
w(i,i') \propto
\frac{\mbox{exp}\left[\theta \left(\bfa_i^{\T}\bfa_{i'} - c\right)\right]}
{1+\mbox{exp}\left[\theta \left(\bfa_i^{\T}\bfa_{i'} - c\right)\right]}.
\label{softweight}
\end{eqnarray}
The ``proportional'' relation ``$\propto$'' in (\ref{softweight})
means the weights $w(i,i')$ are typically normalized to sum to unity, 
i.e., $\sum_{i'=1}^M w(i,i') = 1$ for any given $i$.
The alignment penalty used in (\ref{obj-CB2}) corresponds almost 
everywhere to the special 
and extreme case of $\theta\rightarrow\infty$;
for $\theta < \infty$, 
$w(i,i')$ is a smooth, monotonic function of the number of attributes 
shared by items $i$ and $i'$, rather than an abrupt, step function 
(see Figure~\ref{showstep}).

For $L_{\gAB}$ (\ref{obj-gAB}), the gradient with respect to $\bfp_u$ 
again remains the same,
$\grad^{\gAB}_u = \grad^{\AB}_u = \grad^{\BL}_u$, while the gradient with 
respect to $\bfq_i$ simply becomes
\begin{eqnarray}
\grad^{\gAB}_i &\propto&
\sum_{u \in T_{\cdot i}} -(r_{ui} - \bfp_u^{\T}\bfq_i)\bfp_u + 
 \lambda\gamma
 \left[\bfq_i - \sum_{i'=1}^M w(i,i')\bfq_{i} 
 \right].
\label{dQv2smooth}
\end{eqnarray}
Using smoother weights would allow
all items that share attributes with $i$ to contribute to the shrinkage 
effect, not 
just those that share at least a certain number of attributes with it.
Moreover, their contributions would be adaptive --- the amount of 
``pull'' that item $i'$ exerts on the feature vector of 
item $i$ is appropriated by how many attributes they share in common. 
Depending on how much information there is in the data, this
could potentially enhance the effectiveness of the alignment penalty. 

\subsubsection{A related method: Tag informed CF}
\label{sec:tag}

\def\TG{{\mbox{\rm\tiny TG}}}

Noticing that many commercial recommender engines allow users to create 
personalized tags,
\citet{cf-YZ} proposed a method to exploit information from these tags. 
Following the work of \citet{rel-MF},
their idea was to ``make two user-specific latent 
feature vectors as similar as possible if the two users have similar 
tagging history'' by adding a tag-based penalty to the baseline 
optimization problem:
\[
\underset{\bfP,\bfQ}{\min} \quad 
 L_{\BL}(\bfP,\bfQ) + \underbrace{\lambda
 \sum_{u=1}^N
 \sum_{u'=1}^N \| \bfp_u - \bfp_{u'} \|^2 w(u,u')
 }_{\mbox{tag-based penalty}},
\]
where $w(u,u')$ is a measure of similarity between 
two users based on their tagging history.
Interestingly, if we replace the word ``user'' with 
``item'' and the phrase ``tagging history'' with ``content'' or 
``attributes'', the same idea can be applied to items, i.e.,
\begin{eqnarray}
\underset{\bfP,\bfQ}{\min} \quad L_{\TG}(\bfP,\bfQ) = 
 L_{\BL}(\bfP,\bfQ) + 
 \lambda\gamma
 \sum_{i=1}^M
 \sum_{i'=1}^M \| \bfq_i - \bfq_{i'} \|^2 w(i,i'),
\label{obj-TG}
\end{eqnarray}
where $w(i,i')$ is the similarity between two items based on their 
content information, and the subscript ``TG'' stands for ``tag'' 
indicating where the original idea came from.
But since 
\[
 \|\bfq_i - \bfq_{i'}\|^2 = 
 \|\bfq_i\|^2 + \|\bfq_{i'}\|^2 - 2 \bfq_i^{\T} \bfq_{i'},
\]
it is easy to see that this leads to a similar but slightly 
different mathematical formulation, essentially consisting 
of 
\bitem

\item[(i)] penalizing $\|\bfp_u\|^2$ and $\|\bfq_i\|^2$ by different 
amounts (even if $\gamma=1$) --- in particular, the penalty in front of 
$\|\bfq_i\|^2$ is multiplied by $(1+2w_{i\cdot})$, where $$w_{i\cdot} 
\equiv \sum_{i'=1}^M w(i,i');$$ and

\item[(ii)]
using the generalized version of our alignment penalty 
(\ref{obj-gAB}), up to the specific
choice of $w(i,i')$ itself.
\eitem
Again, for $L_{\TG}$ (\ref{obj-TG}) the gradient with respect to $\bfp_u$ 
remains the same,
$\grad^{\TG}_u = \grad^{\AB}_u = \grad^{\BL}_u$, while
the gradient with respect to $\bfq_i$ becomes
\begin{eqnarray}
\grad^{\TG}_i &\propto&
\sum_{u \in T_{\cdot i}} -(r_{ui} - \bfp_u^{\T}\bfq_i)\bfp_u + \lambda\gamma
\left[ (1+2w_{i\cdot})\bfq_i - 2\sum_{i'=1}^M w(i,i') \bfq_{i'} 
 \right].
\label{dQv3}
\end{eqnarray}
We can see that, when compared with (\ref{dQv2smooth}), 
the selective shrinkage effect is somewhat attenuated in (\ref{dQv3}).
This is most clearly seen if we normalize the weights 
to sum to one, i.e., $w_{i\cdot} = \sum_{i'=1}^M w(i,i') = 1$. 
Then, (\ref{dQv3}) simply becomes 
\begin{eqnarray}
\grad^{\TG}_i &\propto&
\sum_{u \in T_{\cdot i}} -(r_{ui} - \bfp_u^{\T}\bfq_i)\bfp_u + 3\lambda\gamma
\left[ \bfq_i - \frac{2}{3} \sum_{i'=1}^M w(i,i') \bfq_{i'} 
 \right].
\label{dQv3-easy}
\end{eqnarray}
Equation~(\ref{dQv3-easy}) reveals a curious factor of $2/3$ in front of 
the weighted centroid, which clearly dampens 
this algorithm's corresponding shrinkage effect.
 
One of the similarity measures used by \citet{cf-YZ} is the cosine similarity,
\beqn
\label{eq:cosine}
w(i,i') = 
 \frac{\bfa_i^{\T} \bfa_{i'}}
      {\|\bfa_i\|\|\bfa_{i'}\|}.
\eeqn
Although other similarity measures can also be used, 
for binary attributes (see Section~\ref{sec:main}, 
page~\pageref{sec:main}) the cosine 
similarity has an intuitive appeal as it amounts to
something easily interpretable:
\beqn
\label{eq:cosine-meaning}
w(i,i') = 
 \frac{(\mbox{\# attributes shared by $i$ and $i'$})}
      {\sqrt{(\mbox{\# attributes in $i$})(\mbox{\# attributes in $i'$})}}.
\eeqn

\subsection{Regression-constrained factorization}
\label{sec:CB1}

\def\RC{{\mbox{\rm\tiny RC}}}

Another idea for incorporating content information stored in the matrix 
$\bfA$ is to use a regression-style constraint, forcing each item feature 
vector to be a function of the item's content attributes, so that items 
with identical attributes are mapped to the same feature vector. This 
method was first introduced by our group in a short conference paper 
\citep{Forbes11}. 

\subsubsection{Details}
\label{sec:CB1-details}
Specifically, the constraint is
\begin{eqnarray}
\bfQ = \bfA \bfB,
\label{linear-constraint}
\end{eqnarray}
where $\bfB$ is a $D \times K$ matrix. Each {\em column} of $\bfB$ behaves 
like a (vector) regression coefficient that maps the items to a latent 
feature using their content attributes. Each {\em row} of $\bfB$ can be 
viewed as a $K$-dimensional latent feature vector for the corresponding 
attribute.

Under the constraint (\ref{linear-constraint}), the factorization
(\ref{product}) becomes
\[
 \bfR \approx
 \bfP \bfQ^{\T} = \bfP \bfB^{\T} \bfA^{\T}
=
\underbrace{\left[\begin{array}{c} 
\bfp_1^{\T} \\ 
\bfp_2^{\T} \\ 
\vdots \\
\bfp_N^{\T} \\ 
\end{array}\right]}_{N \times K}
\underbrace{\bfB^{\T}}_{K \times D}
\underbrace{\left[\begin{array}{cccc}
\bfa_1 & \bfa_2 & \cdots & \bfa_M
\end{array}\right]}_{D \times M},
\]
and the optimization problem (\ref{obj-BL}) 
becomes
\begin{eqnarray}
\underset{\bfP,\bfB}{\min} \quad
L_{\RC}(\bfP,\bfB) = 
\sum_{(u,i)\in T} (r_{ui} - \bfp_{u}^{\T} \bfB^{\T} \bfa_{i})^2 
+ \lambda\left( \sum_{u} \|\bfp_u \|^2 + 
                \gamma \|\bfB \|^2
         \right).
\label{obj-CB1}
\end{eqnarray}

Again, the
alternating gradient descent algorithm is applicable.
The gradient of $L_{\RC}$ with respect to $\bfp_u$,
$\grad^{\RC}_u$, is the same as $\grad^{\BL}_u$ --- Eq.~(\ref{dQu}), 
except that we replace 
$\bfq_i$ with $\bfB^{\T}\bfa_i$, i.e.,
\begin{eqnarray}
\grad^{\RC}_u &\propto& 
\sum_{i \in T_{u\cdot}} -(r_{ui} - \bfp_u^{\T}\bfB^{\T}\bfa_i)\bfB^{\T}\bfa_i 
+ \lambda\bfp_u.
\end{eqnarray}
Using the fact that $d(\bfx^{\T} \bfM \bfy)/d\bfM = \bfx\bfy^{\T}$,
we can derive easily that the gradient of $L_{\RC}$ with respect to the 
matrix 
$\bfB$ is
\begin{eqnarray}
\grad^{\RC}_{\bfB} &\propto&
\sum_{(u,i) \in T} -(r_{ui} - \bfp_u^{\T}\bfB^{\T}\bfa_i)\bfa_i\bfp_u^{\T} 
+ \lambda\gamma\bfB.
\label{dQB}
\end{eqnarray}
At iteration $(j+1)$, the updating equations are:
\begin{eqnarray}
\bfp_u^{(j+1)} &=& \bfp_u^{(j)} -
 \eta
 \grad^{\RC}_u\left(\bfp_u^{(j)},\bfB^{(j)}\right), \label{newU1} \\
\bfB^{(j+1)} &=& \bfB^{(j)} - 
 \eta
 \grad^{\RC}_{\bfB}\left(\bfp_u^{(j)},\bfB^{(j)}\right). \label{newV1}
\end{eqnarray}

\subsubsection{Related literature}

The idea of incorporating regression relationships into latent factor 
models also has a long history. For example, ecologists used to apply a 
multivariate technique known as correspondence analysis \citep{benzecri, 
greenacre-bk} and fit so-called ordination models to sort species and 
geographical sites with latent coordinates \citep[e.g.,][]{unimode}; sites 
with similar conditions would have close-by coordinates, and 
likewise for species that prefer similar environments. Later, canonical 
correspondence analysis (CCA) was introduced \citep{canoco}, which 
constrains the latent site coordinates to be linear functions of actual 
environmental measurements at those sites. CCA has since become an 
extremely popular technique in the field of environmental ecology 
\citep{unimode-bk}.

\section{Experiments}
\label{sec:expr}

In this section, we describe the data sets we used and the experiments we 
performed to compare and evaluate various content-boosted MF algorithms 
against the baseline MF algorithm. We use the acronyms BL, AB, gAB, TG, 
and RC to refer to the algorithms; these acronyms should be self-evident 
from Sections~\ref{sec:BL} and \ref{sec:main}. Table~\ref{tab:all-methods} 
briefly summarizes all the algorithms being compared and studied.

\begin{table}[ht]
\caption{\label{tab:all-methods}%
Summary of algorithms compared.}
\centering
\fbox{%
\begin{tabular}{l|ll|lccc}
    & & & \multicolumn{4}{c}{other details where applicable}\\
label & \multicolumn{2}{c|}{obj.~func.} 
& $\gamma$ & $c$ & $w(i,i')$ & $\theta$ \\
\hline
BL  & $L_{\BL}$   & Eq.~(\ref{obj-BL})   & $N/M$ & -- & -- & -- \\
AB  & $L_{\AB}$   & Eq.~(\ref{obj-CB2}) & $N/M$ & 1 & -- & -- \\
gAB & $L_{\gAB}$  & Eq.~(\ref{obj-gAB}) & $N/M$ & 1 
	& Eq.~(\ref{softweight})$^{\dagger}$ & 1 \\
TG  & $L_{\TG}$   & Eq.~(\ref{obj-TG})  & $N/(3M)$ & -- 
	& Eq.~(\ref{eq:cosine})$^{\dagger}$  & -- \\ 
RC  & $L_{\RC}$   & Eq.~(\ref{obj-CB1}) & $N/D$ & -- & -- & --     
\end{tabular}}
\begin{footnotesize}
\begin{tabular}{ll}
$^{\dagger}$ & The weights are normalized to sum to one for every $i$.
\end{tabular}
\end{footnotesize}
\end{table}

\subsection{The scaling factor $\gamma$}
\label{sec:gamma-more}

As we have alluded to earlier (Section~\ref{sec:gamma}), the purpose of 
the scaling factor $\gamma$ is to balance the two penalties --- the one on 
$\sum \|\bfp_u\|^2$ and the other on $\sum \|\bfq_i\|^2$ (or $\|\bfB\|^2$ 
in the case of RC) --- so that the objective function is not dominated by 
either the user or the item side of the equation. Since the quantity $\sum 
\|\bfp_u\|^2$ remained constant in this paper and the algorithms differed 
only in terms of how they regularized the $\bfq_i$'s, the use of $\gamma$ 
also allowed us to compare all algorithms on the same scale. With this in 
mind, we used $\gamma = N/M$ for (BL, AB, gAB) and $\gamma=N/D$ for RC. 
For TG, recall that, when the weights were normalized to sum to one, every 
$\|\bfq_i\|^2$ was multiplied by a factor of $1+2w_{i\cdot}=3$ (see 
Section~\ref{sec:tag}). In order to compare everything on the same scale, 
we calibrated this extra factor of $3$ back to $1$ by choosing 
$\gamma=N/(3M)$ for TG.

\subsection{Data sets}
\label{sec:data}

We used two data sets --- ``Recipes'' and ``Movies''.
The data set, ``Recipes'', is a subset of data crawled from 
\url{http://allrecipes.com/} by \citet{Forbes11}, including only recipes 
rated by at least $90$ users, and users who rated at least $50$ recipes. 
The data set, ``Movies'', is the ``MovieLens 100K'' data set from 
\url{http://www.grouplens.org/}. 

\begin{table}[ht]
\centering
\caption{Summary statistics for data sets.}\label{datastats}
\fbox{%
\begin{tabular}{l|r|r}
 & Recipes & Movies\\
\hline
\# of users, $N$			& 1,706 &  943		\\
\# of items, $M$			& 1,040 &  1,682	\\
\# of attributes, $D$			& 1,057 &  19		\\
\# of known ratings, $|T|$		& 64,941 & 100,000 \\
density ratio, $|T|/(MN)$  & 3.7\%$^{\dagger}$ & 6.3\% 
\end{tabular}}\\[1mm]
\begin{footnotesize}
\begin{tabular}{lp{0.475\textwidth}}
$^{\dagger}$ &
Notice that this ratio would have been even lower had we used the full 
recipe data from \cite{Forbes11}.
\end{tabular}
\end{footnotesize}
\end{table}

For ``Recipes'', the ratings are integers between $0$ and $5$, and the 
binary attribute $a_{id}$ is an indicator of whether recipe $i$ contains 
ingredient $d$. For ``Movies'', the ratings are integers between $1$ and 
$5$, and $a_{id}$ is an indicator of whether movie $i$ belongs to genre 
$d$ --- notice that the same movie can (and often do) belong to multiple 
genres. Table~\ref{datastats} contains summary statistics about these two 
data sets.

\subsection{Evaluation}
\label{sec:eval}

To compare and evaluate different algorithms, we repeated the same 
experiment $15$ times. Each time, we sampled 50\% of the user-item pairs 
$(u,i) \in T$ to serve as a hold-out validation set, denoted by $T'$. 
Using the remaining 50\% of the known ratings, we learned the matrices 
$\bfP$ and $\bfQ$ (or $\bfB$ in the case of RC) with different 
algorithms. Ratings for all $(u,i) \in T'$ were predicted by $\wh r_{ui} 
= \bfp_u^{\T}\bfq_i$ (or $\wh r_{ui} = \bfp_u^{\T} \bfB^{\T}\bfa_i$ in the 
case of RC)\footnote{The predicted rating was actually $\wh r_{ui} + \wh 
\mu + \wh \alpha_u + \wh \beta_i$; see Section~\ref{sec:anova}.} --- with 
proper truncation if $\wh r_{ui}$ fell outside $[0,5]$ (for ``Recipes'') 
or $[1,5]$ (for ``Movies'') --- and evaluated by
the mean absolute error (MAE) metric:
$$\mbox{MAE} = \frac{1}{|T'|}\sum_{(u,i)\in T'} |r_{ui}-\wh{r}_{ui}|.$$
Many researchers \citep[e.g.,][]{cf-YZ, ensembleMF} have considered the 
MAE more appropriate for discrete ratings, and the literature is 
increasingly favoring the use of the MAE as opposed to the root mean 
squared error (RMSE), which dominated the Netflix contest. For each 
algorithm, we examined factorizations of a few different 
dimensions, in particular, $K = 5, 10$ and $15$.

\subsection{Additional details for AB and gAB}
\label{sec:CB2-param}

\begin{figure}[ht] 
\centering 
\includegraphics[width=\textwidth]{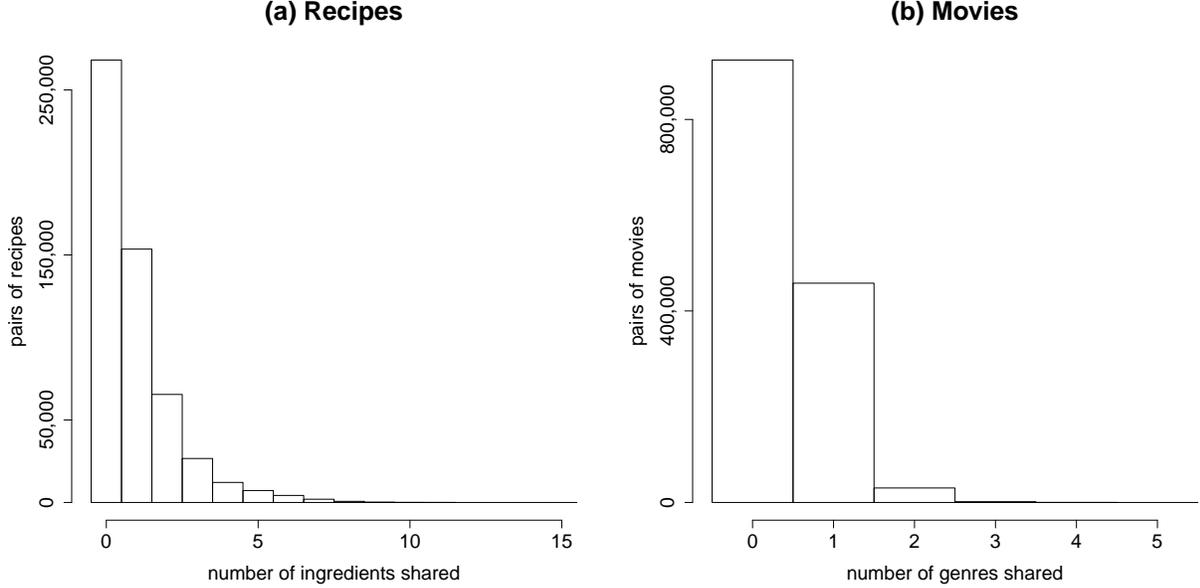} 
\caption{Distribution of the number of attributes shared by pairs of 
items.} 
\label{fig:hist} 
\end{figure}

For both data sets, most items do {\em not} share any attribute in common; 
for those that do, the number of attributes shared is typically small (see 
Figure~\ref{fig:hist}). Thus, we chose $c=1$ for AB and gAB, activating 
the alignment penalty as long as two items shared any attribute at all. 

Generally speaking, one can certainly regard $c$ as an additional tuning 
parameter for AB, but if performance is measured with gross overall 
metrics such as the MAE or the RMSE, then the range of reasonable choices 
for $c$ is fairly limited in our opinion. We think the best strategy is to 
choose $c$ so that the alignment penalty is activated for a certain $x\%$ 
of the item-pairs, and the sensible range for $x$ is somewhere between 
$10$ and $50$. If only a handful of item-pairs were subject to the 
alignment penalty, the overall MAE or RMSE would barely be affected. On 
the other hand, if more than half of the item-pairs were subject to such a 
penalty, items would almost certainly be shrunken blindly toward those 
with which they have little in common. The limited range of sensible 
values for $x$ and the discrete nature of $c$ often greatly restrict the 
choice of $c$. Take the ``Movies'' data set, for example. Choosing $c \geq 
2$ would have resulted in $x \leq 2.2$, whereas choosing $c = 0$ would 
have resulted in $x = 100$ (by definition), so the only sensible choice 
remaining is $c=1$, which gives $x \approx 35$.

As for gAB, it is clear that a large smoothing parameter $\theta$ will 
cause it to behave very much like AB, whereas a small $\theta$ will 
essentially eliminate the effect of the alignment penalty. To focus on 
main ideas rather than fine details, we only provide an {\em illustration} 
of this algorithm using $\theta=1$.

\subsection{Initialization}
\label{sec:init}
\def\svd{{\mbox{\rm\tiny SVD}}}
\def\rand{{\mbox{\rm\tiny RANDOM}}}

We have already mentioned that, when both $\bfP$ and $\bfQ$ are unknown, 
the optimization problem (\ref{obj-BL}) is not convex, which means the 
alternating gradient descent algorithm will give us local solutions at 
best. Hence, a good initialization strategy is useful. 

\subsubsection{SVD strategy}

For given $K$, one way to obtain reasonably good initial values of $\bfP$ 
and $\bfQ$ is as follows. First, impute the missing entries of $\bfR$ with 
predictions from a certain rudimentary model (more below) --- call the 
resulting matrix $\bfR_*$. Then, apply regular SVD and obtain the best 
rank-$K$ approximation to $\bfR_*$:
\[
\bfR_* \approx \bfP_* \bfD_* \bfQ_*^{\T}.
\]
Finally, initialize 
$\bfP$ with $\bfP^{(0)}_{\svd}=\bfP_* \bfD_*^{1/2}$ and 
$\bfQ$ with $\bfQ^{(0)}_{\svd}=\bfQ_* \bfD_*^{1/2}$. In practice, 
since both $\bfP^{(0)}$ and $\bfQ^{(0)}$ are orthogonal matrices, such an 
initialization strategy is often enough to guard against degeneracy even 
though the optimization problem (\ref{obj-BL}) is somewhat ill-posed 
without explicit orthogonality constraints (see Section~\ref{sec:svd}).

The ANOVA model (\ref{eq:anova}) can be used as a rudimentary prediction 
model for imputing the missing entries. But since the ANOVA model was 
actually removed prior to the application of any matrix factorization 
techniques (Section~\ref{sec:anova}), all imputed values should just be 
zero --- this would 
correspond to imputing the missing entries with predictions from the ANOVA 
model before the normalization took place.

It is easy to see that such an initialization strategy would be applicable 
to BL, AB, gAB, and TG. For RC, however, an extra step would be required 
to obtain $\bfB^{(0)}$ from $\bfQ^{(0)}$. Since the RC constraint is 
$\bfQ = \bfA \bfB$, the most natural way to do 
so would be to initialize $\bfB$ with
\beqn
\label{eq:Q2B}
 \bfB^{(0)}_{\svd} = \left(\bfA^{\T} \bfA\right)^{-1} \bfA^{\T} 
 \bfQ^{(0)}_{\svd},
\eeqn
or, if $D > M$ (in which case $\bfA^{\T}\bfA$ would not be invertible),
\beqn
\label{eq:Q2B-ridge}
 \bfB^{(0)}_{\svd} = \left(\bfA^{\T} \bfA + \delta \bfI \right)^{-1} 
 \bfA^{\T} \bfQ^{(0)}_{\svd}
\eeqn
for some $\delta>0$. 
Our default choice was to set $\delta$ to the 
median value of the diagonal elements in $\bfA^{\T}\bfA$.

\subsubsection{Mixed strategy}

\begin{figure}[ht]
\centering
\vspace{-8mm}
\includegraphics[width=0.55\textwidth, angle=270]{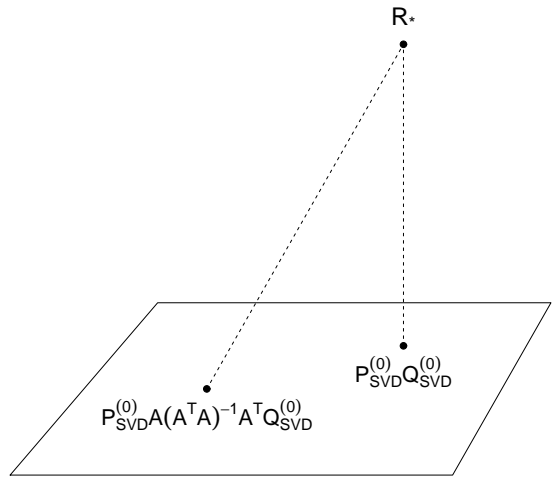}
\vspace{-8mm}
\caption{\label{fig:showproj}%
A geometric explanation of why the SVD initialization strategy forces RC 
into a relative disadvantage. In this illustration, 
$\bfP^{(0)}_{\svd}$ is fixed;
$\bfQ^{(0)}_{\svd}$ gives the best factorization of $\bfR_{*}$; and
anything other than $\bfQ^{(0)}_{\svd}$ gives a worse factorization.}
\end{figure}

While practically useful on its own, the aforementioned SVD strategy 
posed a subtle 
problem for comparison: it forced RC 
into a relative disadvantage. This is because, if
$\langle \bfP^{(0)}_{\svd}, \bfQ^{(0)}_{\svd} \rangle$ is a 
reasonably good initial factorization of $\bfR$, then $\langle 
\bfP^{(0)}_{\svd}, \bfA \bfB^{(0)}_{\svd} \rangle$ will {\em not} be as 
good, since
\[
\bfA \bfB^{(0)}_{\svd} = 
\bfA \left( \bfA^{\T} \bfA \right)^{-1} \bfA^{\T} \bfQ^{(0)}_{\svd} 
\]
is a simply projected version of $\bfQ^{(0)}_{\svd}$.
Figure~\ref{fig:showproj} provides a geometric explanation of why this is 
the case.

For a fair comparison of all algorithms, we therefore used a {\em mixed} 
strategy for initialization. 
More specifically, the matrix $\bfP$ was initialized with
\beqnn
\bfP^{(0)} &=& \kappa \bfP^{(0)}_{\svd} + (1-\kappa) \bfP^{(0)}_{\rand},
\eeqnn
where $\bfP^{(0)}_{\svd}$ was obtained using the SVD strategy, and 
$\bfP^{(0)}_{\rand}$ was a random matrix whose elements were sampled 
independently from $\normal{0}{\sigma^2}$. The same 
procedure was used to initialize $\bfQ$ and/or $\bfB$. 
For given $K$, the parameters $\kappa$ and $\sigma$ were chosen 
separately for (BL, AB, 
gAB, TG) and for RC so that the initial factorizations yielded 
approximately the same level of predictive performance for all the 
algorithms (see Figure~\ref{fig:rslt}).

\subsection{The choice of $\lambda$}
\label{sec:lambda}

Our mixed initialization strategy (Section~\ref{sec:init}), which ensures 
that the initial factorization has approximately the same performance for 
all algorithms, and the way we have scaled the penalty terms 
(Section~\ref{sec:gamma-more}), so that the penalty on $\sum \|\bfq_i\|^2$ 
(or $\sum \|\bfb_d\|^2$ in the case of RC) is on the same order of 
magnitude as the penalty on $\sum \|\bfp_u\|^2$ --- a quantity that 
remains constant for all algorithms, imply that, for the purpose of fair 
comparison, we could (and should) use the same $\lambda$ for all 
algorithms.

Table~\ref{tab:lambda} lists the $\lambda$'s we used for all the 
experiments. 
Our $\lambda$'s increased with $K$, the dimension (or rank) of the 
factorization, because more regularization was needed for factorization 
models that contained more parameters. 
For any given $K$, larger $\lambda$'s were used for 
the ``Movies'' data set than for the ``Recipes'' data set because the 
``Recipes'' data set was more sparse, i.e., the ratio $|T|/N$ was smaller 
(see Table~\ref{datastats}). This meant that, in the case of BL for 
example, the same level of regularization as measured by the ratio,
\[
 \frac{\sum_{(u,i) \in T} (r_{ui} - \bfp^{\T}_u \bfq_i)^2}
      {\lambda(\sum_{u} \|\bfp_u\|^2 + \gamma \sum_{i} \|\bfq_i\|^2)},
\]
could be achieved with a smaller $\lambda$.

\begin{table}[ht]
\centering
\caption{\label{tab:lambda}%
The size of the penalty ($\lambda$) and 
the learning rate ($\eta$) used for different experiments.}
\fbox{
\begin{tabular}{c|cc|cc}
    & \multicolumn{2}{c|}{$\lambda$} 
    & \multicolumn{2}{c}{$\eta$ ($\times 10^{-3}$)} \\
$K$ & Movies & Recipes & Movies & Recipes \\
\hline
5   & 25     & 8  & 2.0  & 2.0 \\
10  & 50     & 12 & 1.0  & 1.5 \\
15  & 75     & 16 & 0.5  & 1.0 \\
\end{tabular}}
\end{table}

\subsection{Convergence criterion and the learning rate $\eta$}

All algorithms were presumed to have reached convergence when the percent 
improvement in their respective objective functions fell below a 
pre-specified threshold, that is, when 
\beqn
\label{eq:convg}
\frac{L^{(j)} - L^{(j+1)}}
     {L^{(j)}} < \varepsilon.
\eeqn
We used $\varepsilon=0.005$ for all algorithms.

For gradient descent algorithms, it is well understood that $\eta$ should 
be kept fairly small to ensure that we are moving in a descent direction 
at each iteration. On the other hand, for practical reasons (e.g., so that 
the algorithm doesn't take forever to finish running) we'd like to use the 
largest $\eta$ feasible --- one that still ensures that we are moving 
downhill. For the convergence criterion (\ref{eq:convg}), however, it was 
critical that the learning rate $\eta$ did not differ significantly for 
different algorithms. Suppose algorithm 1 used a relatively large $\eta$ 
and algorithm 2 used a relatively small one. Then, {\em relative to} 
algorithm 1, algorithm 2 could ``converge'' prematurely according to 
(\ref{eq:convg}) simply because the small $\eta$ did not allow its 
objective function to change very much from iteration to iteration. 
Therefore, for any given $K$, not only did we use the same $\lambda$ for 
all algorithms, we also used the same $\eta$ (see Table~\ref{tab:lambda}).

\subsection{Results}
\label{sec:rslt}

\begin{figure}[hp]
\centering
\includegraphics[width=0.385\textwidth, angle=0]{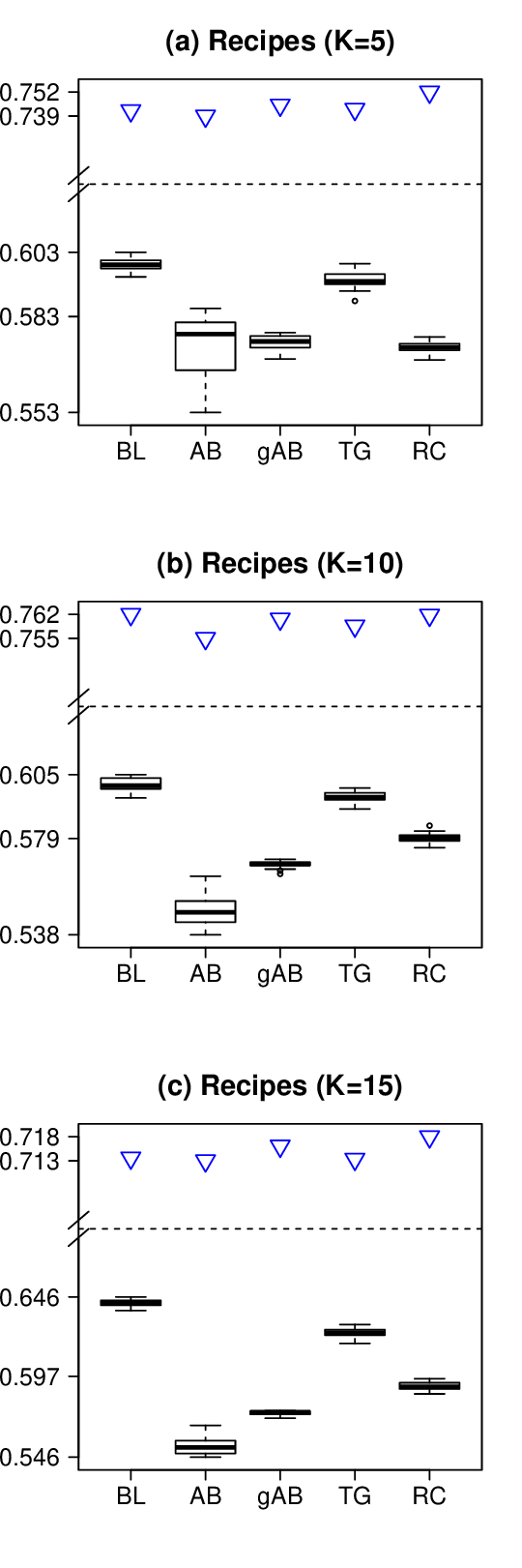}
\includegraphics[width=0.385\textwidth, angle=0]{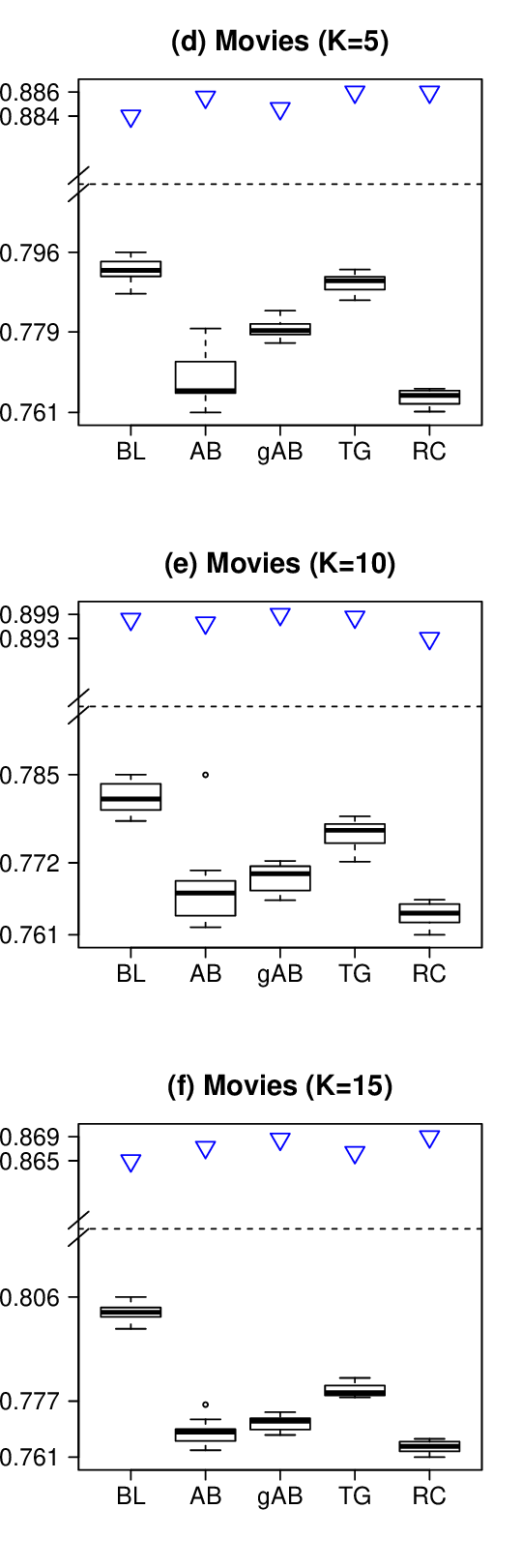}
\caption{\label{fig:rslt}%
Mean absolute errors (MAEs) on hold-out validation sets from 15 repeated 
runs. For each run, the data set was randomly split into a training set 
and a validation set (see Section~\ref{sec:eval}). The inverted triangles 
($\bigtriangledown$) on the top indicate the average MAEs on the 
validation set using the initial values for each respective algorithm, 
i.e., $\bfP^{(0)}$ and $\bfQ^{(0)}$ (or $\bfB^{(0)}$ in the case of RC). 
These are shown here to emphasize the fact that all algorithms were 
started with initial values of approximately the same quality, so that our 
overall comparison is fair (see Section~\ref{sec:init}). Notice the broken 
vertical axes.}
\end{figure}

Figure~\ref{fig:rslt} summarizes our experimental results. We can see 
that, starting with initial factorizations of roughly the same quality and 
using the same level of regularization (as controlled by $\gamma$ and 
$\lambda$), the same learning rate ($\eta$), and the same convergence 
criterion (\ref{eq:convg}), the content-boosted algorithms (AB, gAB, TG, 
RC) generally had lower MAEs than the baseline algorithm (BL). The 
performance of TG appears to trail behind that of similar algorithms in 
the same class (i.e., AB, gAB). We think this is due to its much 
dampened shrinkage effect (Section~\ref{sec:tag}).

\section{Discussions}
\label{sec:disc}

We now describe useful by-products from these content-boosted 
matrix-factorization techniques.

\subsection{More interpretable recommendations}
\label{sec:CB2-rslt}

By explicitly pulling ``similar'' items together in the latent feature 
space, where ``similarity'' is defined by the contents of the items, the 
alignment-biased algorithms (AB, gAB, TG) produce recommendations that are 
easier to explain. Research has shown that the ``why'' dimension of 
recommendation --- the ability ``to reason to the user why certain 
recommendations are presented'' \cite{recsys11-ebay} --- improves the 
effectiveness of the recommender system, especially as measured by the 
conversion rate \cite{rec-transparency}.

\begin{figure}[hp]
\centering
\includegraphics[width=\textwidth]{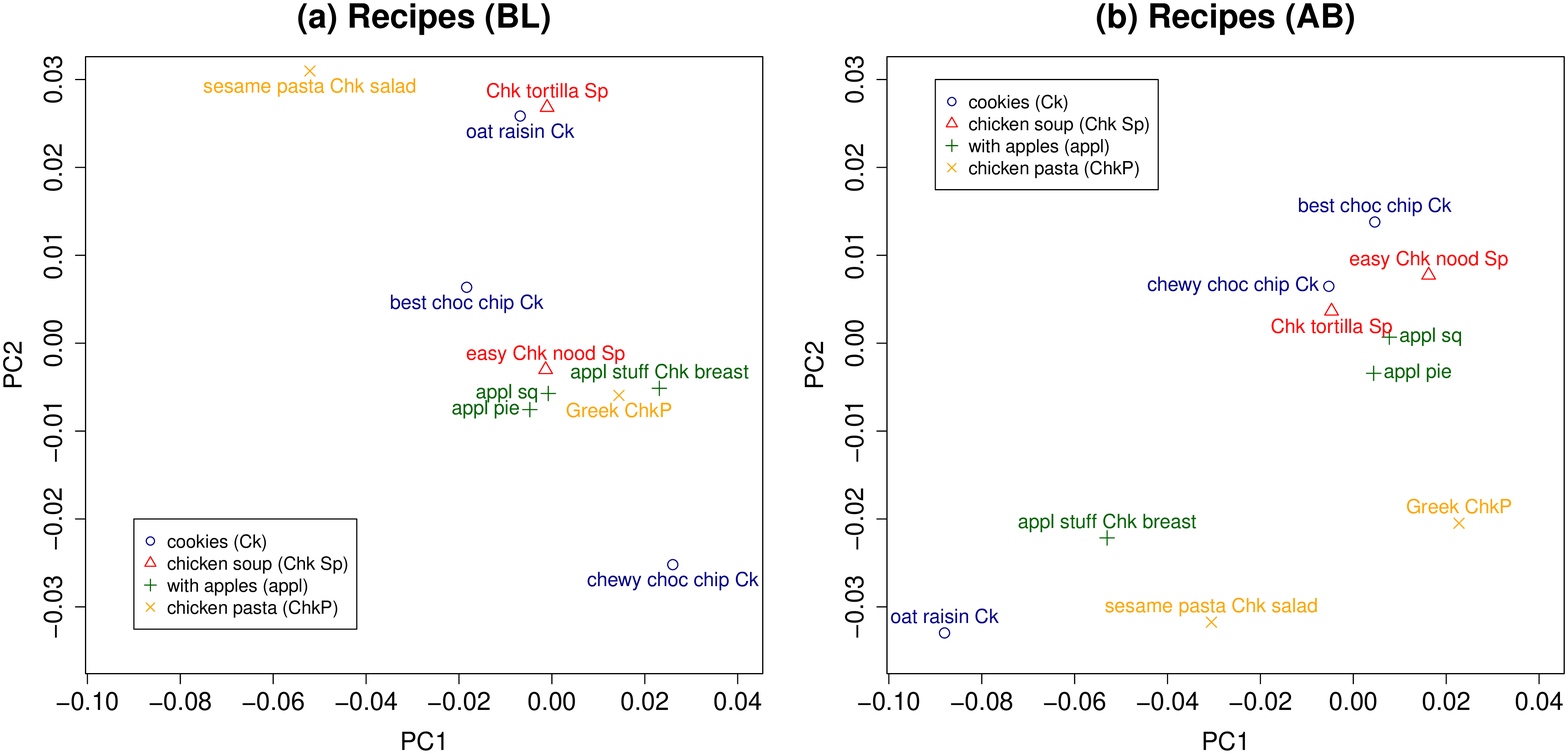} \\
\includegraphics[width=\textwidth]{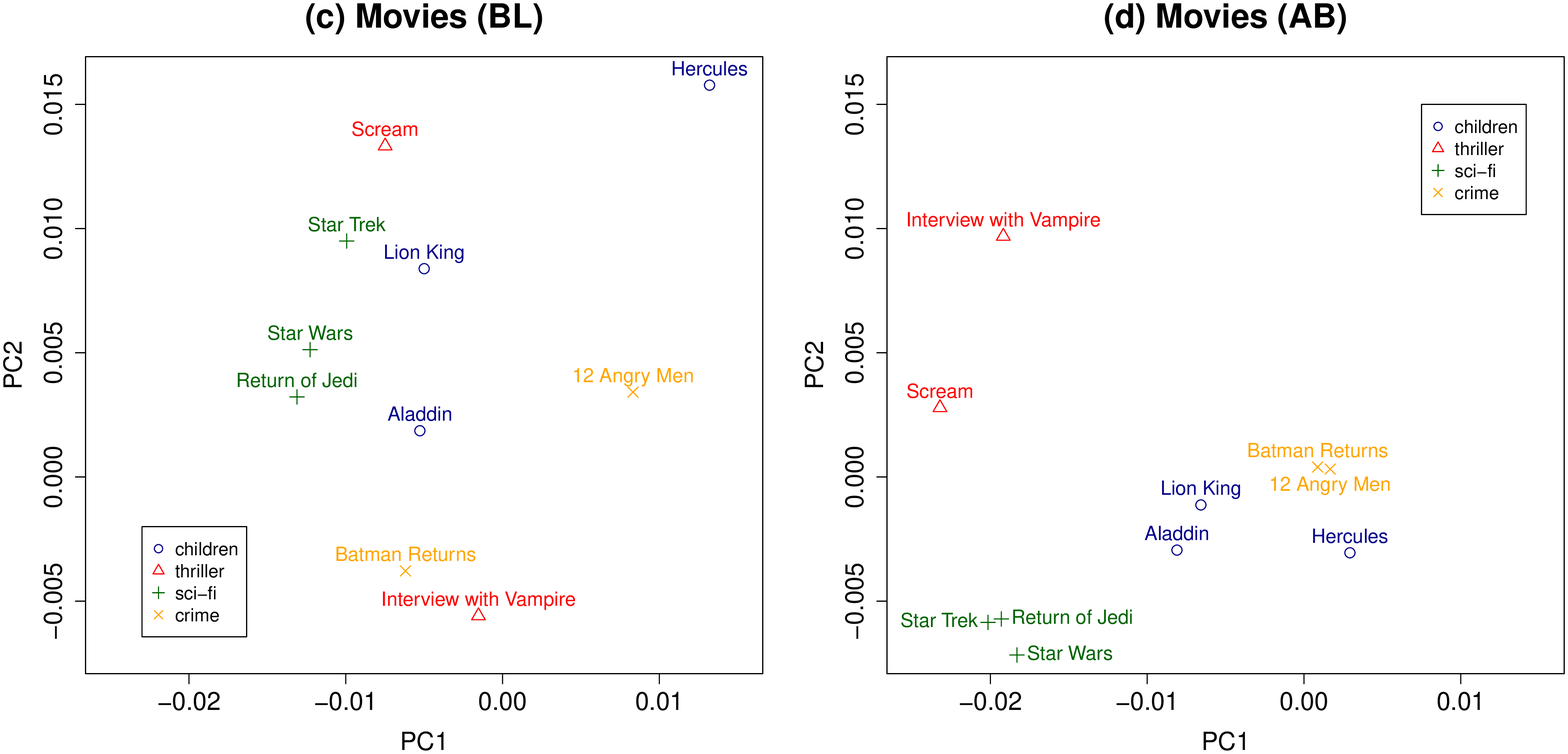}
\caption{\label{map}%
Feature vectors for selected items --- $5$-dimensional matrix 
factorization solutions projected onto 2 leading principal components for 
2D-display. BL = ``baseline'' algorithm (Section~\ref{sec:BL}); AB = 
``alignment-biased'' algorithm (Section~\ref{sec:CB2}).}
\end{figure}

To illustrate, we selected a number of movies from a few distinct genres 
(e.g., thriller, sci-fi), as well as a number of recipes from a few 
different categories (e.g., soup, pasta, cookie), and plotted their latent 
feature vectors $\bfq_i \in \mathbb{R}^5$ from BL and from AB, using the 
first two principal components (Figure~\ref{map}). Here, we chose to 
illustrate the $5$-dimensional solutions because showing 
higher-dimensional solutions in 2D would have created more distortion.

As expected, recipes containing common ingredients --- e.g., ``Greek 
chicken pasta'' and ``sesame paste chicken salad'' --- have been pulled 
closer together by the alignment-biased algorithm. The two chicken soups 
are closer to each other. The dish, ``apple stuffed chicken breast'', is 
now closer to chicken pastas than to apple deserts. On the other hand, 
``oatmeal raisin cookies'' are pulled away from the other two, 
``chocolate-chip cookies'' because the key ingredients are different. 

Likewise, movies belonging to the same genres are now closer to each 
other, e.g., ``Interview with Vampire'' and ``Scream'' --- both thrillers. 
The same can be said about the three children's movies and the three 
science fictions. Clearly, the coordinate maps produced by the 
alignment-biased algorithm, AB, are much easier to explain to consumers.

\subsection{Measure of content similarity}
\label{sec:CB1-rslt}

The regression-constrained algorithm (RC) allows us to compute the 
similarity of two content attributes, $d$ and $d'$, using their latent 
feature vectors, e.g.,
\begin{eqnarray}
\label{cosine}
 \cos(d, d') &=& \frac{\bfb_d^{\T}\bfb_{d'}}
  {\|\bfb_d\|\|\bfb_{d'}\|},
\end{eqnarray}
where $\bfb_d^{\T}$ is the $d$-th row of the matrix $\bfB$. Notice that, 
as a measure of similarity, (\ref{cosine}) is {\em not} based on the 
simple notion of 
co-occurrence --- merely counting how often two attributes are shared by 
the same item, since $\bfb_d$ is driven by both content attributes 
and user preferences. 

Table~\ref{cos} shows a few examples from both data sets, for pairs of 
ingredients and genres ranging from being highly similar ($\cos \gg 0$) to 
being highly dissimilar ($\cos \ll 0$). All results in this table are 
based on $K=15$. It is well-known that high-dimensional vectors are more 
likely to be orthogonal ($\cos \approx 0$) than low-dimensional ones. 
We chose to calculate (\ref{cosine}) using 
relatively high-dimensional feature vectors so that cosine-values far away 
from zero were more meaningful.

Some of these pairs are not too surprising. For example, it is easy to see 
that people who like ``Thai chili sauce'' would also like ``jalapeno 
peppers'' (Table~\ref{cos}a) --- both spicy ingredients. Likewise, we are 
hardly amazed that those who like ``crime'' movies will probably also like 
``horror'' movies, and that the genre ``children'' goes much better with 
``adventure'' than with ``documentary'' ($\cos \approx 0.74 > 0 
\mbox{~vs.~} \cos \approx -0.52 < 0$; Table~\ref{cos}b).

Other pairs, however, are much less obvious. For example, 
Table~\ref{cos}(b) shows that users who like ``war'' movies are more 
likely to favor ``animation'' movies over ``action'' movies ($\cos \approx 
0.34 > 0 \mbox{~vs.~} \cos \approx -0.21 < 0$). Similarly, 
Table~\ref{cos}(a) 
tells us that users who like ``smoked ham'' will probably also like 
``chocolate mint wafer candy'' and that, if a user likes ``cottage 
cheese'', he or she may detest ``Swiss cheese''. This kind of insight 
about the contents is a unique by-product of the regression-constrained 
algorithm, and some of these novel insights can be commercially useful. 
For example, Table~\ref{cos}(a) suggests that ``firm tofu'' might be used 
to replace ``mozzarella'' in some recipes --- if you are familiar with 
both ingredients, you may very well appreciate that this is not a bad idea 
at all.

\begin{table}[ht]
\centering
\caption{Selected pairs of attributes and their cosine
similarity (\ref{cosine}) based on their latent feature 
vectors in $\mathbb{R}^{15}$.}\label{cos}
(a) Recipes\\[2mm]
\fbox{%
\begin{tabular}{p{0.4\textwidth}|p{0.4\textwidth}|r}
Ingredient 1 & Ingredient 2 & Cosine \\
\hline
Thai chili sauce or hot sauce &	jalapeno peppers & 0.9395 \\
chocolate mint wafer candy	& smoked ham &	0.9070 \\
mozzarella &	firm tofu	& 0.8828 \\
can jellied cranberry sauce	& ginger garlic paste &	0.5283 \\
almonds &	pork sausage &	 0.0001 \\
bread crumbs &	black olive &	-0.0001 \\
cottage cheese & Swiss cheese &	-0.5017 \\
can corn &	golden delicious apple & -0.8709 \\
can sweetened condensed milk &	seedless green grapes	& -0.8944 \\
can beef broth & dry sherry &	-0.9402
\end{tabular}}\\[5mm]
(b) Movies\\[2mm]
\fbox{%
\begin{tabular}{p{0.4\textwidth}|p{0.4\textwidth}|r}
Genre 1 & Genre 2 & Cosine \\
\hline
adventure	& children	& 0.7421 \\
crime	        & horror	& 0.6456 \\
action	        & sci-fi	& 0.4354 \\
animation	& war	        & 0.3431 \\
documentary	& musical	& 0.0129 \\
comedy	        & film-noir	& -0.0160 \\
action	        & war	        & -0.2078 \\
comedy	        & mystery	& -0.4348 \\
children        & documentary	& -0.5234 \\
action	        & drama	        & -0.8973 
\end{tabular}}
\end{table}

\section{Future work}
\label{sec:future}

Before we end, we would like to briefly mention some open problems for 
future research.

\subsection{Matrix completion}
\label{sec:matcompletion}

The CF problem can also be formulated as a {\em matrix completion} problem 
\cite{matcomplete}:
\beqn
 \min_{\wh\bfR} &\quad& \mbox{rank}(\wh\bfR), \label{eq:minrank}\\
 \mbox{s.t.}    &\quad& \wh{r}_{ui} = r_{ui} \quad\mbox{for}\quad (u,i) \in T. \label{eq:completion}
\eeqn
That is, we'd like to fill in the missing entries of $\bfR$ while keeping 
the rank of the completed matrix $\wh\bfR$ as low as possible. The 
rationale behind rank minimization is similar to that behind the matrix 
factorization approach: we believe that user-preferences are driven by 
only a few key factors; therefore, the rank of the rating matrix cannot be 
very high. The two approaches thus share a common philosophical 
underpinning, but they differ in that the matrix factorization approach is 
more explicit about the nature of the low-rankness.

Suppose $\sigma_1(\wh\bfR) \geq \sigma_2(\wh\bfR) \geq ... \geq 
\sigma_{\min(N,M)}(\wh\bfR) \geq 0$ are the (ordered) singular values of 
$\wh\bfR$. Then,
\beqn
\mbox{rank}(\wh\bfR) = \sum_{j=1}^{\min(N,M)} I(\sigma_j(\wh\bfR) \neq 0), \label{eq:rank-l0}
\eeqn
and the rank-minimization problem 
(\ref{eq:minrank})-(\ref{eq:completion}), like all problems having to do 
with minimizing the number of non-zero elements (e.g., variable selection 
\cite{lasso}, compressed sensing \cite{compressed-sensing}), is NP-hard. 
However, (\ref{eq:rank-l0}) shows that $\mbox{rank}(\wh\bfR)$ is 
equivalent to the $l_0$-norm of the vector, $\boldsigma=(\sigma_1, 
\sigma_2, ..., \sigma_{\min(N,M)})^{\T}$. It has been well established 
\cite[e.g.,][and references therein]{donoho-l1, 
candes-tao-signal-recovery-IEEE, candes-tao-signal-recovery-Comm} that the 
$l_1$-norm is a good convex relaxation of the $l_0$-norm so, instead of 
(\ref{eq:minrank}), one can consider
\beqn
 \min_{\wh\bfR} \quad \|\wh\bfR\|_{*}  \label{eq:min-nuclear}
\eeqn
where $\|\wh\bfR\|_{*}$ denotes the {\em nuclear-norm} (also known as the {\em trace-norm}) of $\wh\bfR$, defined as
\[
 \|\wh\bfR\|_{*} = \sum_{j=1}^{\min(N,M)} \sigma_j(\wh\bfR),
\]
or the $l_1$-norm of $\boldsigma$. 

The nuclear-norm minimization problem 
(\ref{eq:min-nuclear})-(\ref{eq:completion}) is convex, and can be solved 
by semi-definite programming (SDP). Remarkably, under certain conditions, 
e.g., restricted isometry \cite{rip, rip-general} or matrix incoherence 
\cite{matcomplete}, the convex optimization problem involving the 
nuclear-norm ($l_1$-norm minimization) can be shown to actually solve the 
minimum-rank problem ($l_0$-norm minimization). This is an extraordinary 
set of achievements, made possible by some of the best mathematicians in 
the world, and it is the main reason why the matrix completion approach is 
gaining much attention as of late.

Notwithstanding its strong mathematical foundations, the matrix completion 
approach has some limitations, too. For example, \citet{matcomplete-limit} 
showed examples where one could fill in a sparse matrix in different ways, 
while maintaining the same (low) rank for the completed matrix. 
Computationally, SDP is still only realistic for fairly small matrices, 
which is why the matrix completion approach has not yet been widely 
applied to the CF problem, although much research is being devoted to it. 

Most importantly for us, however, it is less clear how to generalize the 
mathematical problem (\ref{eq:min-nuclear})-(\ref{eq:completion}) so as to 
bring in extra content information --- notice that, in our approach, 
content-based constraints and penalties were introduced to operate 
directly (and conveniently) on the latent feature vectors, but the matrix 
completion approach does not admit an explicit parameterization to 
facilitate the kind of extensions we have proposed. Thus, it appears that, 
as far as the matrix completion problem is concerned, a different paradigm 
would be needed for incorporating content information.

\subsection{Generative models}
\label{sec:simulations}

Readers may have noticed the apparent lack of simulation experiments in 
our study. Simulation experiments are, in fact, rare in the CF literature. 
We think this is due to the lack of widely accepted {\em generative 
models} for user-rating data. While it is certainly possible to use the 
latent factor model (\ref{product}) itself to simulate the ratings, this 
clearly would favor the matrix factorization approach in any subsequent 
performance comparison, and it is easy to understand why the CF community 
has not found such an approach to be terribly interesting or informative. 
For our study, the question of what makes a suitable {\em generative 
model} is even more complex, since we consider extra content information. 
Additional research is clearly needed in order to address these issues.

\section{Summary}
\label{sec:summary}

To sum up, we have focused on different ways to incorporate content 
information directly into the matrix-factorization approach for 
collaborative filtering. Our methodology consists of imposing either an 
``alignment penalty'' (Section~\ref{sec:CB2}), effectively shrinking items 
that share common attributes toward each other, or a regression-style 
constraint (Section~\ref{sec:CB1}), forcing the latent item-features to be 
functions of content attributes. Experiments with two data sets have shown 
that these content-boosted algorithms can not only achieve better 
recommendation accuracy, they can also produce novel, commercially useful 
insights about the contents themselves, as well as more interpretable 
recommendations.

Our treatment of the problem is by no means thorough. For example, it is 
certainly possible to envision different types of penalties and 
constraints, and we have not yet attempted to study the theoretical 
properties of these different approaches. As we mentioned in 
Section~\ref{sec:future}, it would be interesting to think about how to 
bring in content information for the matrix completion (as opposed to the 
matrix factorization) approach, and it would be useful to come up with 
plausible generative model for describing user-rating data. This is a rich 
area with many opportunities for continued research. We hope that our 
paper has not only outlined a few useful ideas for practitioners, but also 
made it easier for researchers to think about this type of problems in a 
more systematic manner.

\subsection*{Acknowledgments}

This research was partially supported by the Natural Sciences and 
Engineering Research Council (NSERC) of Canada, and by the University of 
Waterloo (UW). The first author did most of her work on this project while 
an undergraduate student at UW. The authors thank Peter Forbes for 
allowing them to use the recipe data that he assembled, and Laura Ye for 
creating a denser subset from Peter's original data. The authors also 
thank three referees and an associate editor for their insightful comments 
and constructive criticism. They are especially grateful to referee No.~3 
for introducing them to the related TG algorithm (Section~\ref{sec:tag}).

\bibliographystyle{/u/m3zhu/unsrtnat}
\bibliography{references}

\begin{thebibliography}{44}
\providecommand{\natexlab}[1]{#1}
\providecommand{\url}[1]{\texttt{#1}}
\expandafter\ifx\csname urlstyle\endcsname\relax
  \providecommand{\doi}[1]{doi: #1}\else
  \providecommand{\doi}{doi: \begingroup \urlstyle{rm}\Url}\fi

\bibitem[Feuerverger et~al.(2012)Feuerverger, He, and
  Khatri]{netflix-review-statsci}
A.~Feuerverger, Y.~He, and S.~Khatri.
\newblock Statistical significance of the {N}etflix challenge.
\newblock \emph{Statistical Science}, 27\penalty0 (2):\penalty0 202--231, 2012.

\bibitem[Salakhutdinov et~al.(2007)Salakhutdinov, Mnih, and Hinton]{cf-rbm}
R.~R. Salakhutdinov, A.~Mnih, and G.~Hinton.
\newblock Restricted {B}oltzmann machines for collaborative filtering.
\newblock In \emph{Proceedings of the 24th International Conference on Machine
  Learning}, pages 791--798, 2007.

\bibitem[Koren(2008)]{netflix-nbr}
Y.~Koren.
\newblock Factorization meets the neighborhood: A multifaceted collaborative
  filtering model.
\newblock In \emph{Proceedings of the 14th {ACM SIGKDD} International
  Conference on Knowledge Discovery and Data Mining}, pages 426--434, 2008.

\bibitem[Koren et~al.(2009)Koren, Bell, and Volinsky]{Koren09}
Y.~Koren, R.~Bell, and C.~Volinsky.
\newblock Matrix factorization techniques for recommender systems.
\newblock \emph{Computer}, 42\penalty0 (8):\penalty0 30--37, 2009.

\bibitem[Su and Khoshgoftaar(2009)]{survey-CF}
X.~Su and T.~M. Khoshgoftaar.
\newblock A survey of collaborative filtering techniques.
\newblock \emph{Advances in Artificial Intelligence}, 2009, 2009.
\newblock Article {ID} 421425.

\bibitem[Park et~al.(2006)Park, Pennock, Madani, Good, and DeCoste]{Park06}
S.~T. Park, D.~Pennock, O.~Madani, N.~Good, and D.~DeCoste.
\newblock Na\"{i}ve filterbots for robust cold-start recommendations.
\newblock In \emph{Proceedings of the 12th {ACM SIGKDD} International
  Conference on Knowledge Discovery and Data Mining}, pages 699--705, 2006.

\bibitem[Zhao et~al.(2011)Zhao, Feng, Li, and Liu]{Zhao11}
Y.~Zhao, X.~Feng, J.~Li, and B.~Liu.
\newblock Shared collaborative filtering.
\newblock In \emph{Proceedings of the 5th {ACM} Conference on Recommender
  Systems}, pages 29--36, 2011.

\bibitem[Goldberg et~al.(2001)Goldberg, Roeder, Gupta, and Perkins]{Goldberg01}
K.~Goldberg, T.~Roeder, D.~Gupta, and C.~Perkins.
\newblock Eigentaste: A constant time collaborative filtering algorithm.
\newblock \emph{Information Retrieval}, 4:\penalty0 133--151, 2001.

\bibitem[Nguyen et~al.(2007)Nguyen, Denos, and Berrut]{Nguyen07}
A.~T. Nguyen, N.~Denos, and C.~Berrut.
\newblock Improving new user recommendations with rule-based induction on cold
  user data.
\newblock In \emph{Proceedings of the 2007 {ACM} Conference on Recommender
  Systems}, pages 121--128, 2007.

\bibitem[Melville et~al.(2002)Melville, Mooney, and Nagarajan]{Melville02}
P.~Melville, R.~J. Mooney, and R.~Nagarajan.
\newblock Content-boosted collaborative filtering for improved recommendation.
\newblock In \emph{Proceedings of the 18th National Conference on Artificial
  Intelligence}, pages 187--192, 2002.

\bibitem[Semeraro et~al.(2009)Semeraro, Lops, Basile, and
  {de~Gemmis}]{Semeraro09}
G.~Semeraro, P.~Lops, P.~Basile, and M.~{de~Gemmis}.
\newblock Knowledge infusion into content-based recommender systems.
\newblock In \emph{Proceedings of the 3rd {ACM} Conference on Recommender
  Systems}, pages 301--304, 2009.

\bibitem[Zhen et~al.(2009)Zhen, Li, and Yeung]{cf-YZ}
Y.~Zhen, W.-J. Li, and D.-Y. Yeung.
\newblock {TagiCoFi}: Tag informed collaborative filtering.
\newblock In \emph{Proceedings of the 3rd {ACM} Conference on Recommender
  Systems}, pages 69--76, 2009.

\bibitem[Nunes(2009)]{Nunes09}
M.~A.~S.~N. Nunes.
\newblock \emph{Recommender Systems Based on Personality Traits: Could human
  psychological aspects influence the computer decision-making process?}
\newblock VDM Verlag, Berlin, 2009.

\bibitem[Hu and Pu(2011)]{Hu11}
R.~Hu and P.~Pu.
\newblock Enhancing collaborative filtering systems with personality
  information.
\newblock In \emph{Proceedings of the 5th {ACM} Conference on Recommender
  Systems}, pages 197--204, 2011.

\bibitem[Jamali and Ester(2010)]{Jamali10}
M.~Jamali and M.~Ester.
\newblock A matrix factorization technique with trust propagation for
  recommendation in social networks.
\newblock In \emph{Proceedings of the 4th {ACM} Conference on Recommender
  Systems}, pages 135--142, 2010.

\bibitem[Yu et~al.(2011)Yu, Pan, and Li]{Yu11}
L.~Yu, R.~Pan, and Z.~Li.
\newblock Adaptive social similarities for recommender systems.
\newblock In \emph{Proceedings of the 5th {ACM} Conference on Recommender
  Systems}, pages 257--260, 2011.

\bibitem[Katz et~al.(2011)Katz, Ofek, Shapira, Rokach, and Shani]{Katz11}
G.~Katz, N.~Ofek, B.~Shapira, L.~Rokach, and G.~Shani.
\newblock Using {W}ikipedia to boost collaborative filtering techniques.
\newblock In \emph{Proceedings of the 5th {ACM} Conference on Recommender
  Systems}, pages 285--288, 2011.

\bibitem[Forbes and Zhu(2011)]{Forbes11}
P.~Forbes and M.~Zhu.
\newblock Content-boosted matrix factorization for recommender systems:
  Experiments with recipe recommendation.
\newblock In \emph{Proceedings of the 5th {ACM} Conference on Recommender
  Systems}, pages 261--264, 2011.

\bibitem[Sundaresan(2011)]{recsys11-ebay}
N.~Sundaresan.
\newblock Recommender systems at the long tail.
\newblock In \emph{Proceedings of the 5th {ACM} Conference on Recommender
  Systems}, pages 1--5, 2011.

\bibitem[Sinha and Swearingen(2002)]{rec-transparency}
R.~Sinha and K.~Swearingen.
\newblock The role of transparency in recommender systems.
\newblock In \emph{{CHI} '02 Extended Abstracts on Human Factors in Computing
  Systems}, pages 830--831, 2002.

\bibitem[Van~Buskirk(2009)]{netflix-wired-news}
E.~Van~Buskirk.
\newblock How the {Netflix} prize was won, 2009.
\newblock {\it Wired} magazine online,
  \url{http://www.wired.com/business/2009/09/how-the-netflix-prize-was-won/}.

\bibitem[Mardia et~al.(1979)Mardia, Kent, and Bibby]{mardia}
K.~V. Mardia, J.~T. Kent, and J.~M. Bibby.
\newblock \emph{{Multivariate Analysis}}.
\newblock Academic Press, 1979.

\bibitem[Salakhutdinov and Mnih(2008)]{MFprob}
R.~R. Salakhutdinov and A.~Mnih.
\newblock Probabilistic matrix factorizations.
\newblock In J.~Platt, D.~Koller, Y.~Singer, and S.~Roweis, editors,
  \emph{Advances in Neural Information Processing Systems}, volume~20, pages
  1257--1264, 2008.

\bibitem[Funk(2006)]{funk}
S.~Funk.
\newblock {N}etflix update: Try this at home (online blog), 2006.
\newblock \url{http://sifter.org/~simon/journal/20061211.html}.

\bibitem[Volinsky(2010)]{svd-volinsky}
C.~Volinsky.
\newblock Email communications, 28 {A}ugust and 3 {S}eptember, 2010.

\bibitem[Lee and Seung(1999)]{nmf-leeseung}
D.~D. Lee and H.~S. Seung.
\newblock Learning the parts of objects by non-negative matrix factorization.
\newblock \emph{Nature}, 401\penalty0 (6755):\penalty0 788--791, 1999.

\bibitem[Brunet et~al.(2004)Brunet, Tamayo, Golub, and Mesirov]{nmf-metagene}
J.-P. Brunet, P.~Tamayo, T.~R. Golub, and J.~P. Mesirov.
\newblock Metagenes and molecular pattern discovery using matrix factorization.
\newblock \emph{Proceedings of the National Academy of Sciences of the USA},
  101\penalty0 (12):\penalty0 4164--4169, 2004.

\bibitem[Gu et~al.(2010)Gu, Zhou, and Ding]{cf-YZ-NMF}
Q.~Gu, J.~Zhou, and C.~Ding.
\newblock Collaborative filtering: Weighted nonnegative matrix factorization
  incorporating user and item graphs.
\newblock In \emph{Proceedings of the 10th {SIAM} International Conference on
  Data Mining}, pages 199--210, 2010.

\bibitem[Wu(2007)]{ensembleMF}
M.~Wu.
\newblock Collaborative filtering via ensembles of matrix factorizations.
\newblock In \emph{Proceedings of {KDD} Cup and Workshop}, pages 43--47, 2007.

\bibitem[Li and Yeung(2009)]{rel-MF}
W.-J. Li and D.-Y. Yeung.
\newblock Relation regularized matrix factorization.
\newblock In \emph{Proceedings of the 21st International Joint Conference on
  Artificial Intelligence}, pages 1126--1131, 2009.

\bibitem[Benz\'{e}cri(1973)]{benzecri}
J.-P. Benz\'{e}cri.
\newblock \emph{L'Analyse des Donn\'{e}es (Vol.~II): L'Analyse des
  Correspondances}.
\newblock Dunod, Paris, 1973.

\bibitem[Greenacre(1983)]{greenacre-bk}
M.~Greenacre.
\newblock \emph{Theory and Applications of Correspondence Analysis}.
\newblock Academic Press, London, 1983.

\bibitem[ter Braak(1985)]{unimode}
C.~J.~F. ter Braak.
\newblock Correspondence analysis of incidence and abundance data: Properties
  in terms of a unimodal response model.
\newblock \emph{Biometrics}, 41:\penalty0 859--873, 1985.

\bibitem[ter Braak(1986)]{canoco}
C.~J.~F. ter Braak.
\newblock Canonical correspondence analysis: A new eigenvector technique for
  multivariate direct gradient analysis.
\newblock \emph{Ecology}, 67\penalty0 (5):\penalty0 1167--1179, 1986.

\bibitem[ter Braak(1996)]{unimode-bk}
C.~J.~F. ter Braak.
\newblock \emph{Unimodal Models to Relate Species to Environment}.
\newblock DLO-Agricultural Mathematics Group, Wageningen, 1996.

\bibitem[Cand\`{e}s and Recht(2009)]{matcomplete}
E.~J. Cand\`{e}s and B.~Recht.
\newblock Exact matrix completion via convex optimization.
\newblock \emph{Foundations of Computational Mathematics}, 9:\penalty0
  717--772, 2009.

\bibitem[Tibshirani(1996)]{lasso}
R.~Tibshirani.
\newblock Regression shrinkage and selection via the {L}asso.
\newblock \emph{Journal of the Royal Statistical Society Series B},
  58:\penalty0 267--288, 1996.

\bibitem[Donoho(2006{\natexlab{a}})]{compressed-sensing}
D.~L. Donoho.
\newblock Compressed sensing.
\newblock \emph{{IEEE} Transaction on Information Theory}, 52\penalty0
  (4):\penalty0 1289--1306, 2006{\natexlab{a}}.

\bibitem[Donoho(2006{\natexlab{b}})]{donoho-l1}
D.~L. Donoho.
\newblock For most large underdetermined systems of linear equations the
  minimal $l_1$-norm solution is also the sparsest solution.
\newblock \emph{Communications on Pure and Applied mathematics}, 59\penalty0
  (6):\penalty0 797–--829, 2006{\natexlab{b}}.

\bibitem[Cand\`{e}s and Tao(2006)]{candes-tao-signal-recovery-IEEE}
E.~J. Cand\`{e}s and T.~Tao.
\newblock Near-optimal signal recovery from random projections: Universal
  encoding strategies?
\newblock \emph{{IEEE} Transaction on Information Theory}, 52\penalty0
  (12):\penalty0 5406--5425, 2006.

\bibitem[Cand\`{e}s et~al.(2006)Cand\`{e}s, Romberg, and
  Tao]{candes-tao-signal-recovery-Comm}
E.~J. Cand\`{e}s, J.~K. Romberg, and T.~Tao.
\newblock Stable signal recovery from incomplete and inaccurate measurements.
\newblock \emph{Communications on Pure and Applied Mathematics}, 59\penalty0
  (8):\penalty0 1207–--1223, 2006.

\bibitem[Cand\`{e}s and Tao(2005)]{rip}
E.~J. Cand\`{e}s and T.~Tao.
\newblock Decoding by linear programming.
\newblock \emph{{IEEE} Transaction on Information Theory}, 51\penalty0
  (12):\penalty0 4203–--4215, 2005.

\bibitem[Recht et~al.(2010)Recht, Fazel, and Parrilo]{rip-general}
B.~Recht, M.~Fazel, and P.~A. Parrilo.
\newblock Guaranteed minimum-rank solutions of linear matrix equations via
  nuclear norm minimization.
\newblock \emph{{SIAM} Review}, 52:\penalty0 471–--501, 2010.

\bibitem[Shi and Yu(2010)]{matcomplete-limit}
X.~Shi and P.~S. Yu.
\newblock Limitations of matrix completion via trace norm minimization.
\newblock \emph{{SIGKDD} Explorations}, 12\penalty0 (2):\penalty0 16--20, 2010.

\end{thebibliography}

\end{document}